\documentclass[10pt,twocolumn,letterpaper]{article}

\usepackage[pagenumbers]{wacv} 

\usepackage{graphicx}
\usepackage{amsmath}
\usepackage{amssymb}
\usepackage{booktabs}

\usepackage{booktabs}
\usepackage{flushend}
\usepackage{bigstrut}
\usepackage{multirow}

\usepackage{algorithm}
\usepackage{algpseudocode}
\usepackage{bbding}

\usepackage[accsupp]{axessibility}  

\def\eg{\textit{e.g.}}
\def\ie{\textit{i.e.}}
\def\etal{\textit{et al.}}

\usepackage{color}
\definecolor{lightblue}{RGB}{0, 110, 204}

\usepackage[pagebackref=true,breaklinks=true,letterpaper=true,citecolor=lightblue,colorlinks,bookmarks=false]{hyperref}

\usepackage[capitalize]{cleveref}
\crefname{section}{Sec.}{Secs.}
\Crefname{section}{Section}{Sections}
\Crefname{table}{Table}{Tables}
\crefname{table}{Tab.}{Tabs.}

\def\wacvPaperID{27} 
\def\confName{WACV}
\def\confYear{2024}

\begin{document}

\title{Efficient Transferability Assessment for Selection of Pre-trained Detectors}

\author{
Zhao Wang$^1$ \quad Aoxue Li$^2$\thanks{Corresponding author.} \quad Zhenguo Li$^2$ \quad Qi Dou$^1$\\
$^1$The Chinese University of Hong Kong \quad $^2$Huawei Noah’s Ark Lab\\
{\tt \small \{zwang21@cse., qidou@\}cuhk.edu.hk \quad lax@pku.edu.cn \quad li.zhenguo@huawei.com}
}


\maketitle

\begin{abstract}
Large-scale pre-training followed by downstream fine-tuning is an effective solution for transferring deep-learning-based models. Since finetuning all possible pre-trained models is computational costly, we aim to predict the transferability performance of these pre-trained models in a computational efficient manner. Different from previous work that seek out suitable models for downstream classification and segmentation tasks, this paper studies the efficient transferability assessment of pre-trained object detectors. To this end, we build up a detector transferability benchmark which contains a large and diverse zoo of pre-trained detectors with various architectures, source datasets and training schemes. Given this zoo, we adopt 7 target datasets from 5 diverse domains as the downstream target tasks for evaluation. Further, we propose to assess classification and regression sub-tasks simultaneously in a unified framework. Additionally, we design a complementary metric for evaluating tasks with varying objects. Experimental results demonstrate that our method outperforms other state-of-the-art approaches in assessing transferability under different target domains while efficiently reducing wall-clock time 32$\times$ and requires a mere 5.2\% memory footprint compared to brute-force fine-tuning of all pre-trained detectors. 
\end{abstract}

\section{Introduction}

\begin{figure}[t]
  \centering
   \includegraphics[width=\linewidth]{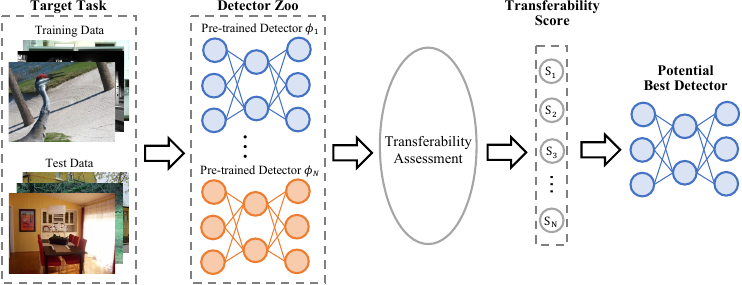}
   \vspace{-7mm}
   \caption{Illustration for selection of pre-trained detectors. The selection is performed by efficient transferability assessment.}
   \label{fig:teaser}
\end{figure}

Under a paradigm of large-scale model pre-training \cite{he2016deep, huang2017densely, tan2019efficientnet, wang2020deep, he2020momentum, chen2020simple, grill2020bootstrap, chen2021exploring} and downstream fine-tuning \cite{guo2019spottune, wortsman2022robust, chen2020adversarial}, starting from a good pre-trained model is crucial.
Nevertheless, it is too costly to perform selection of pre-trained models by brute-forcibly fine-tuning all available pre-trained models on a given downstream task \cite{yosinski2014transferable, kornblith2019better}.
Fortunately, existing works have shown the advantages to efficiently evaluate the transferability of pre-trained models with specific design for image classification \cite{nguyen2020leep, li2021ranking, you2021logme, you2022ranking, shao2022not, ding2022pactran, pandy2022transferability} and semantic segmentation \cite{pandy2022transferability, agostinelli2022transferability} tasks without fine-tuning.
They usually estimate the transferability by measuring the class separation of representations extracted by different pre-trained models \cite{nguyen2020leep, ding2022pactran, pandy2022transferability, agostinelli2022transferability}. 
While previous works consider estimating the transferability of classification or segmentation task, this paper aims to rank the transferablity of pretrained models for object detection, as shown in Figure \ref{fig:teaser}. Since object detection methods address both classification and regression sub-tasks together in the same scheme, most assessment methods based on class-separation can hardly be applied, especially for the single-class object detection.

To evaluate transferability assessment of object detection, we build up a challenging yet practical experimental setup with a zoo of diverse pretrained detectors. Specifically, we collect $33$ large-scale pre-trained object detectors including two-stage \cite{ren2015faster, cai2018cascade}, single-stage \cite{lin2017focal, sun2021sparse}, and transformer \cite{zhu2020deformable, dai2021up} based detection architectures.
Meanwhile, they are equipped with different backbones, varying in ResNets \cite{he2016deep}, ResNeXts \cite{xie2017aggregated}, and RegNets \cite{radosavovic2020designing} and pre-trained with various datasets \cite{lin2014microsoft,kuznetsova2020open,deng2009imagenet}.
Moreover, we adopt $6$ downstream tasks from $5$ diverse domains under different scenarios including general objects \cite{everingham2010pascal}, driving \cite{cordts2016cityscapes, han2021soda10m}, dense prediction \cite{shao2018crowdhuman}, Unmanned Aerial Vehicle (UAV) \cite{zhu2021detection} and even medical lesions \cite{yan2018deeplesion}. In contrast, previous classification assementment methods usually simultaneously rank over 10$\sim$20 pretrained models and test on datasets from at most 3 domains \cite{nguyen2020leep, ding2022pactran, pandy2022transferability}. Upon the detection transferability benchmark, we propose to assess classification and regression sub-tasks simultaneously in a unified framework. Moreover, a complementary metric is designed for better assess tasks with various object scales. 

Our main contributions are summarized as:
\begin{itemize}
    \item This paper studies a crucial yet underexplored problem: efficient transferability assessment of pre-trained detectors. 
    \item We build up a challenging detection transferability benchmark, containing $33$ pre-trained detectors with different architectures, different source datasets and training schemes, and evaluating on diverse downstream target tasks. A series of metrics are designed to effectively assess these pre-trained detectors. 
    \item Extensive experimental results upon this challenging transferability benchmark show the effectiveness and robustness of the proposed assessment framework compared with previous SOTA methods.  Most importantly, it achieves over $32\times$ wall-clock time speedup and only $5.2\%$ memory footprint requirement compared with brute-force fine-tuning.
\end{itemize}

\section{Related Work}


\subsection{Transferability of Pre-trained Models}
Assessing the transferability of pre-trained models is an essential and crucial task.
Early works \cite{yosinski2014transferable, kornblith2019better} have studied the transferability of the neural networks based on various layers within a single model and various models within a model zoo while the conclusions were drawn from too expensive fine-tuning (see Sec. \ref{sec:efficiency}), which is not affordable.
Further, the transferability of deep knowledge was studied by evaluating the task relatedness \cite{zamir2018taskonomy, dwivedi2019representation, tran2019transferability, tan2021otce} and building attribution graphs \cite{song2019deep, song2020depara}.
But these solutions still need costly downstream training and they are not applicable when meeting multi-scale features, multiple sub-tasks, and diverse detection heads, which are the key components in object detection.

Recently, several papers introduced efficient transferability metrics for classification.
LEEP \cite{nguyen2020leep} proposed to efficiently evaluate the transferability of pre-trained models for supervised classification tasks by calculating the log expectation of the empirical predictor.
However, the performance of LEEP degrades when the number of classes within the source data is less than that of the target task. A series of following works \cite{li2021ranking, agostinelli2022transferability} further improve LEEP for multi-class classification and pixel-level classification (i.e., semantic segmentation). $\mathcal{N}$-LEEP \cite{li2021ranking} tried to improve LEEP with Principal Component Analysis  on the model outputs.
and established a neural checkpoint ranking benchmark for classification.
To better assess multi-class classification, Ding \etal \cite{ding2022pactran} proposed a series of Cross Entropy (CE) based metrics for pre-trained classification models.
P\'andy \etal \cite{pandy2022transferability} proposed to estimate the pairwise Gaussian class separability using the Bhattacharyya coefficient, which could be applicable for both classification and segmentation.
In SFDA \cite{shao2022not}, the authors aimed to leverage the fine-tuning dynamics into transferability measurement in a self-challenging fisher space, which degraded the efficiency.

Different from previous works designed for image-level or pixel-level classification, this paper aims to tackle more challenging pre-trained detector assessment. We thus build up a detector transferability benchmark, which simultaneously ranks 33 pre-trained models over 6 target datasets from 5 diverse domains, comparing with $10\!\!\!\!\sim\!\!\!\!20$ pre-trained models and 3 target domains in classification scenarios. 
Moreover, assessing pre-trained detectors should consider both classification and regression sub-tasks together.   
LogME \cite{you2021logme, you2022ranking} first extends the transferability assessment from classification to regression and thus can be used as a baseline to assess the transferability of pre-trained detectors. However, it is designed for general regression task, without considering the multi-scale characteristics and inherent relation between coordinates in bounding box regression sub-task. To address these issues, we extend LogME to a series of metrics with special design for object detection.




\subsection{Object Detection}

Object detection is a practical computer vision task that aims to detect the objects and recognize the corresponding classes from an input image.
The object detectors are always trained under supervised \cite{ren2015faster, cai2018cascade, lin2017focal, tian2019fcos, zhu2020deformable} and self-supervised \cite{wei2021aligning, yang2021instance, dai2021up, bar2022detreg} schemes on the large-scale datasets \cite{lin2014microsoft, kuznetsova2020open, gupta2019lvis, shao2019objects365, deng2009imagenet}.
The supervised detectors are trained with ground truth bounding boxes and class labels while the self-supervised ones can only access the training images.
Regarding the design of different detection architectures, there are three main streams, including two-stage \cite{ren2015faster, cai2018cascade, zhang2020dynamic}, single-stage \cite{lin2017focal, tian2019fcos, sun2021sparse}, and transformer \cite{zhu2020deformable, dai2021up, bar2022detreg} based detectors.
A typical two-stage detector \cite{ren2015faster} works with $1$st-stage proposal generator and $2$nd stage bounding box refinement and class recognition while a single-stage detector \cite{lin2017focal} aims to perform dense predictions for the object location and class.
Recent transformer based end-to-end detectors \cite{carion2020end, zhu2020deformable} consider the object detection task as a direct set prediction problem optimized with Hungarian algorithm \cite{kuhn1955hungarian}, in which lots of human-designed complex components are removed.
With these large-scale pre-trained object detectors, how to figure out a good one for a given downstream detection task is crucial but underexplored. 
In this work, we aim to tackle the efficient transferability assessment for pre-trained detectors, which infers the true fine-tuning performance on a give downstream task.

\newcommand{\tabincell}[2]{
\begin{tabular}{@{}#1@{}}#2\end{tabular}
}

\section{Detector Transferability Benchmark}
\label{sec:benchmark}

In this work, we construct a zoo of different detectors pre-trained with various source datasets and  and thus build up a transferability benchmark to measure different assessment metrics. In what follows, we will provide details of the detector transferability benchmark. 

\paragraph{Problem Setup.}
With a given downstream detection task and a detection model zoo consisting of $N$ pre-trained models $\{\mathcal{F}_n\}_{n=1}^N$, the aim of model transferability assessment is to produce a transferability score for every pre-trained detector, and then find the best one for further fine-tuning according to the score ranking.

\begin{table}[t]
  \caption{Pre-training schemes, source datasets and detection architectures used in this work. We include two-stage, single-stage, and transformer based detectors to build a pre-trained detector zoo. These detectors are equipped with different backbones, \eg, ResNets \cite{he2016deep}, ResNeXts \cite{xie2017aggregated}, and RegNets \cite{radosavovic2020designing}.}
  \label{tab:source_models}%
\vspace{-3mm}
  \centering
\scalebox{0.68}{

\begin{tabular}{l|l|lll}
\hline
Scheme & Dataset & Type  & Detector & Backbone \bigstrut\\
\hline
\multirow{29}[22]{*}{Supervised} & \multirow{27}[18]{*}{COCO \cite{lin2014microsoft}} & \multirow{17}[10]{*}{two-stage} & \multirow{4}[2]{*}{FRCNN \cite{ren2015faster}} & R50 \cite{he2016deep} \bigstrut[t]\\
      &       &       &       & R101 \cite{he2016deep} \\
      &       &       &       & X101-32x4d \cite{he2016deep} \\
      &       &       &       & X101-64x4d \cite{he2016deep} \bigstrut[b]\\
\cline{4-5}      &       &       & \multirow{4}[2]{*}{\tabincell{l}{Cascade\\RCNN} \cite{cai2018cascade}} & R50 \cite{he2016deep} \bigstrut[t]\\
      &       &       &       & R101 \cite{he2016deep} \\
      &       &       &       & X101-32x4d \cite{he2016deep} \\
      &       &       &       & X101-64x4d \cite{he2016deep} \bigstrut[b]\\
\cline{4-5}      &       &       & \tabincell{l}{Dynamic\\RCNN} \cite{zhang2020dynamic} & R50 \cite{he2016deep} \bigstrut\\
\cline{4-5}      &       &       & \multirow{5}[2]{*}{RegNet \cite{radosavovic2020designing}} & 400MF \cite{radosavovic2020designing} \bigstrut[t]\\
      &       &       &       & 800MF \cite{radosavovic2020designing} \\
      &       &       &       & 1.6GF \cite{radosavovic2020designing} \\
      &       &       &       & 3.2GF \cite{radosavovic2020designing} \\
      &       &       &       & 4GF \cite{radosavovic2020designing} \bigstrut[b]\\
\cline{4-5}      &       &       & \multirow{3}[2]{*}{DCN \cite{dai2017deformable}} & R50 \cite{he2016deep} \bigstrut[t]\\
      &       &       &       & R101 \cite{he2016deep} \\
      &       &       &       & X101-32x4d \cite{he2016deep} \bigstrut[b]\\
\cline{3-5}      &       & \multirow{9}[6]{*}{single-stage} & \multirow{2}[2]{*}{FCOS \cite{tian2019fcos}} & R50 \cite{he2016deep} \bigstrut[t]\\
      &       &       &       & R101 \cite{he2016deep} \bigstrut[b]\\
\cline{4-5}      &       &       & \multirow{5}[2]{*}{RetinaNet \cite{lin2017focal}} & R18 \cite{he2016deep} \bigstrut[t]\\
      &       &       &       & R50 \cite{he2016deep} \\
      &       &       &       & R101 \cite{he2016deep} \\
      &       &       &       & X101-32x4d \cite{he2016deep} \\
      &       &       &       & X101-64x4d \cite{he2016deep} \bigstrut[b]\\
\cline{4-5}      &       &       & \multirow{2}[2]{*}{\tabincell{l}{Sparse\\ RCNN} \cite{sun2021sparse}} & R50 \cite{he2016deep} \bigstrut[t]\\
      &       &       &       & R101 \cite{he2016deep} \bigstrut[b]\\
\cline{3-5}      &       & transformer & DDETR \cite{zhu2020deformable} & R50 \cite{he2016deep} \bigstrut\\
\cline{2-5}      & \multirow{2}[4]{*}{\tabincell{l}{Open \\Images} \cite{kuznetsova2020open}} & two-stage & FRCNN \cite{ren2015faster} & R50 \cite{he2016deep} \bigstrut\\
\cline{3-5}      &       & single-stage & RetinaNet \cite{lin2017focal} & R50 \cite{he2016deep} \bigstrut\\
\hline
\multirow{4}[4]{*}{\tabincell{l}{Self-\\Supervised}} & \multirow{4}[4]{*}{ImageNet \cite{deng2009imagenet}} & \multirow{2}[2]{*}{two-stage} & SoCo \cite{wei2021aligning} & R50 \cite{he2016deep} \bigstrut[t]\\
      &       &       & InsLoc \cite{yang2021instance} & R50 \cite{he2016deep} \bigstrut[b]\\
\cline{3-5}      &       & \multirow{2}[2]{*}{transformer} & UP-DETR \cite{dai2021up} & R50 \cite{he2016deep} \bigstrut[t]\\
      &       &       & DETReg \cite{bar2022detreg} & R50 \cite{he2016deep} \bigstrut[b]\\
\hline
\end{tabular}%

}

\end{table}%

\paragraph{Pre-trained Source Detectors.}
In the past few years, a great number of detection models arise with very smart design.
Typically, these detectors are trained under supervised \cite{ren2015faster, tian2019fcos, zhu2020deformable} or self-supervised \cite{wei2021aligning, dai2021up} scenarios.
Supervised detectors are always trained and evaluated on large-scale detection datasets, such as COCO \cite{lin2014microsoft} and Open Images \cite{kuznetsova2020open}, while self-supervised ones are trained based on ImageNet \cite{deng2009imagenet}.
In this work, we take three datasets, \ie, COCO, Open Images, and ImageNet, as the source datasets for different pre-training schemes.
Upon these source datasets and training schemes, we use $13$ detection architectures, including two-stage \cite{ren2015faster, cai2018cascade, zhang2020dynamic, radosavovic2020designing, dai2017deformable, wei2021aligning, yang2021instance}, single-stage \cite{lin2017focal, tian2019fcos, sun2021sparse}, and transformer \cite{zhu2020deformable, dai2021up, bar2022detreg} based detectors, for pre-training and obtain a model zoo composed of $33$ various pre-trained detectors.
These detectors are equipped with different backbones, \eg, ResNets \cite{he2016deep}, ResNeXts \cite{xie2017aggregated}, and RegNets \cite{radosavovic2020designing}. The detailed information is shown in Table \ref{tab:source_models}.
With a large variety of pre-trained source detectors, it can be ensured that at least one good model exists for a given target task and our evaluation metric can distinguish between bad and good source models.

\paragraph{Target Datasets.}

The image domain plays an important role for deep-learning-based models, which directly determines the model performance in a transfer learning scenario \cite{pan2009survey, mensink2021factors}.
So we cover a wide range of image domains as a challenging but practical setting with diverse scenarios, including general objects \cite{everingham2010pascal}, driving \cite{cordts2016cityscapes, han2021soda10m}, dense prediction \cite{shao2018crowdhuman}, Unmanned Aerial Vehicle (UAV) \cite{zhu2021detection}, and even medical lesions \cite{yan2018deeplesion}. The detailed information of these target datasets are summarized in Table \ref{tab:target_datasets}.

\begin{table}[t]
  \caption{Target downstream datasets used in this work, in which they are from $5$ diverse domains.}
  \label{tab:target_datasets}%
  \centering
  \vspace{-3mm}
\scalebox{0.8}{

\begin{tabular}{llcc}
\toprule
Dataset & Domain & Classes & Images \\
\midrule
Pascal VOC \cite{everingham2010pascal} & General & 20    & 21K \\
CityScapes \cite{cordts2016cityscapes} & Driving & 8     & 5K \\
SODA \cite{han2021soda10m} & Driving & 6     & 20K \\
CrowdHuman \cite{shao2018crowdhuman} & Dense & 1     & 24K \\
VisDrone \cite{zhu2021detection} & UAV   & 11    & 9K \\
DeepLesion \cite{yan2018deeplesion} & Medical & 8     & 10K \\
\bottomrule
\end{tabular}%

}
\end{table}%

\begin{figure*}[t]
  \centering
   \includegraphics[width=\linewidth]{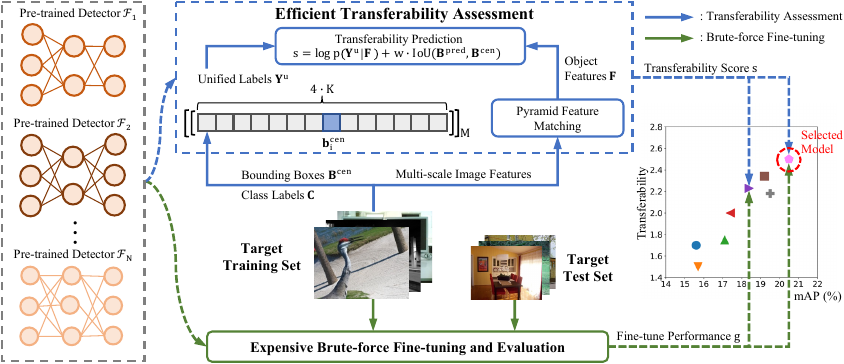}
   \vspace{-7mm}
   \caption{The overview of efficient transferability assessment framework for pre-trained detection models. We build a challenging setting contains various pre-trained object detectors. Based on this challenging setting, we design a pyramid feature matching scheme to handle objects with various sizes and expand the bounding box matrix $\boldsymbol{B}^{cen}$ according to class label matrix $\boldsymbol{C}$ to the unified label matrix $\boldsymbol{Y}^u$ for evaluation. We estimate the maximum evidence $p(\boldsymbol{Y}^u|\boldsymbol{F})$, which indicates the compatibility between the object features $\boldsymbol{F}$ and unified labels $\boldsymbol{Y}^u$. Further, considering IoU as an important metric in object detection, we supply IoU between the predicted bounding boxes and ground truth ones as a complementary term for transferability assessment of detection model.}
   \label{fig:framework}
\end{figure*}

\paragraph{Evaluation Protocol.}

Given pre-trained object detectors $\{\mathcal{F}_n\}_{n=1}^N$ and a downstream dataset $\mathcal{D}_t$, a transferability assessment method will produce the transferability scores $\{s_n\}_{n=1}^N$.
Following the previous works \cite{li2021ranking, you2021logme, agostinelli2022transferability, pandy2022transferability, shao2022not, ding2022pactran, you2022ranking}, we take the fine-tuning performance of pre-trained detectors $\{g_n\}_{n=1}^N$ as the ground truth, \ie, mean Average Precision (mAP) as the metric in object detection.
Ideally, the transferability scores are positively correlated with true fine-tuning performance.
That is, if a pre-trained model $\mathcal{F}_n$ has higher detection mAP than $\mathcal{F}_m$ after fine-tuning ($g_n \textgreater g_m$), the transferability score of $\mathcal{F}_n$ is also expected to be larger than $\mathcal{F}_m$ ($s_n \textgreater s_m$).
So the effectiveness of a transferability metric for assessing the pre-trained detectors is evaluated by the ranking correlation between the ground truth fine-tuning performance $\{g_n\}_{n=1}^N$ and estimated transferability scores $\{s_n\}_{n=1}^N$.
We use \emph{Weighted Kendall’s $\tau_w$} \cite{kendall1938new} as the evaluation metric.
Larger $\tau_w$ indicates better ranking correlation between $\{g_n\}_{n=1}^N$ and $\{s_n\}_{n=1}^N$ and better transferability metric.
$\tau_w$ is interpreted by
\vspace{-2mm}
\begin{equation}
\small
    \tau_w=\frac{2}{N(N-1)}\!\!\sum_{1 \leq n<m \leq N} \!\!\operatorname{sgn}\left(g_n-g_m\right) \operatorname{sgn}\left(s_n-s_m\right).
\label{equ:tw}
\end{equation}
\vspace{-4mm}


\noindent Here $\tau_w$ ranges in $[-1, 1]$, and the probability of $g_n \textgreater g_m$ is $\frac{\tau_w + 1}{2}$ when $s_n \textgreater s_m$.
Moreover, we use Top-1 Relative Accuracy (Rel@1) \cite{li2021ranking} to measure how close model with the highest transferability performs, in terms of fine-tuning performance, compared to the highest performing model.

\section{Detection Model Transferability Metrics}

In this section, we study the problem of efficient transferability assessment for detection model. 
Given a model zoo with a number of pre-trained detectors and a target downstream task, our goal is to predict their transferability performance on the target task efficiently, without brute-force fine-tuning all pre-trained models.
LogME is a classification assessment method designed from the viewpoint of regression \cite{you2021logme}. Thus, it can be used to assess transferability of pre-trained detectors which contain both classification and regression subtasks. However, it is designed for general regression and might fail when tackling the bounding box regression task with multi-scale characteristics of inputs and inherent relation between coordinates in outputs. To this end, as shown in Figure \ref{fig:framework}, we extend LogME to detection scenario by designing a unified framework (U-LogME in Sec. \ref{sec:u_logme}) which assesses multiple sub-tasks and multi-scale features simultaneously. Furthermore, we propose a complementary metric (i.e., IoU-LogME in Sec. \ref{sec:iou_logme}) for better transferability assessment over objects with varying scales. 

\subsection{LogME as a Basic Metric}
\label{sec:logme}
\label{par:logme}

Different from most existing assessment methods that measure class separation of visual features, LogME addresses the problem from the viewpoint of regression. Specifically, it uses a set of Bayesian linear models to fit the features extracted by the pre-trained models and the corresponding labels. The marginalized likelihood of these linear models is used to rank pre-trained models. Since it can address both classification and regression tasks, we can easily extend it to assess object detection framework. 
 
Specifically, we extract multi-scale object-level features of ground-truth bounding boxes by using pre-trained detectors' backbone followed by an ROIAlign layer \cite{ren2015faster}. In this way, for a given pre-trained detector and a downstream task, we can collect the object-level features of downstream task by using the detector and form a feature matrix $\boldsymbol{F}$, with each row $\boldsymbol{f}_i$ denotes an object-level feature vector. For each $\boldsymbol{f}_i$, we also collect its 4-d coordinates of ground-truth bounding box $\boldsymbol{b}_i$ and class label $c_i$ to form a bounding box matrix $\boldsymbol{B}$ and a class label matrix $\boldsymbol{C}$.

For the bounding box regression sub-task, LogME measures the transferability by using the maximum evidence $p(\boldsymbol{B}|\boldsymbol{F})\!\!=\!\!\int p(\boldsymbol{\theta}|\alpha)p(\boldsymbol{B}|\boldsymbol{F},\beta,\boldsymbol{\theta}) d\boldsymbol{\theta}$, where $\boldsymbol{\theta}$ is the parameter of linear model. $\alpha$ denotes the parameter of prior distribution of $\boldsymbol{\theta}$, and $\beta$ denotes the parameter of posterior distribution of each observation $p(\boldsymbol{b}_i|\boldsymbol{f}_i,\beta,\boldsymbol{\theta})$. By using the evidence theory \cite{knuth2015bayesian} and basic principles in graphical models \cite{koller2009probabilistic}, the transferability metric can be formulated as
\begin{equation}
\small
\begin{aligned}
    \operatorname{LogME} =
    &\log p(\boldsymbol{B}|\boldsymbol{F}) \\
    =&\frac{M}{2} \log \beta+\frac{D}{2} \log \alpha-\frac{M}{2} \log 2 \pi \\
    &-\frac{\beta}{2}\|\boldsymbol{F} \boldsymbol{m}-\boldsymbol{B}\|_2^2-\frac{\alpha}{2} \boldsymbol{m}^T \boldsymbol{m}-\frac{1}{2} \log |A|,
\end{aligned}
\label{equ:logme}
\end{equation}
where $\boldsymbol{m}$ is the solution of $\boldsymbol{\theta}$, $M$ is the number of objects, $D$ is the dimension of features, and $A$ is the $L_2$-norm of $\boldsymbol{F}$.

LogME for classification sub-task can be computed by replacing $\boldsymbol{B}$ in Eq. \eqref{equ:logme} with converted one-hot class label matrix. 
By combining the evidences of two sub-tasks together, we obtain a final evidence for ranking the pre-trained detector. 
However, LogME still struggles to rank the pre-trained detectors accurately due to the following four challenges: 1) The single-scale features of LogME is not compatible with the multi-scale features extracted by pyramid network architecture of pre-trained detector. 2) The huge coordinates variances of different-scale objects make it hard to fit by simple linear model used in LogME. 3) The assessment branch of classification sub-task might fail when the downstream task is single-class detection.
4) The mean squared error (MSE) used in LogME isn't scale-invariant and measures each coordinate separately, which is not suitable for bounding box regression. In what follows, we propose how to address these issues. 




\subsection{U-LogME}
\label{sec:u_logme}
This subsection proposes to unify multi-scale features and multiple tasks into an assessment framework. 
Here, we propose a pyramid feature mapping scheme to extract suitable features for different scale objects. 
Meanwhile, we normalize coordinates of bounding boxes and jointly evaluate 4-d coordinates with the same linear network. 
This can help to reduce coordinate variances and thus benefit ranking. 
Furthermore, we merge the class labels and normalized bounding box coordinates into a final ground-truth label to joint assess two sub-tasks with the same model. 
In this way, both single-class and multiple-class downstream detection tasks are unified into one assessment framework. 
We provide more technical details in the following.  

\paragraph{Pyramid Feature Matching.}
Classical object detectors always include a Feature Pyramid Network (FPN) \cite{lin2017feature} like architecture with different feature levels.
The object features are obtained from different levels according to the corresponding object sizes during training, \eg, very small and large objects will be mapped to bottom-level and top-level image features, respectively.
Regarding this, we introduce \emph{pyramid feature matching} to help objects with different sizes find their matched level features.
Following FPN \cite{lin2017feature}, we assign an object to the feature pyramid level $P_l$ by the following:
\vspace{-2mm}
\begin{equation}
\small
    l=\left\lfloor l_0+\log _2(\sqrt{w h} / 224)\right\rfloor,
    \label{equ:pfm}
\end{equation}
where $w$ and $h$ is the width and length of an object on the input image to the network, respectively.
Here $224$ is the ImageNet \cite{deng2009imagenet} training size, and $l_0$ is the feature level mapped by an object with $w\!\times\!h\!=\!224^2$.
Inspired by FCOS \cite{tian2019fcos}, to better handle too small and large objects, objects satisfying $\max(w, h) \textless 64$ and $\max(w, h) \textgreater 512$ are further forcibly assigned to the lowest and highest level of feature pyramid. Given a bounding box and its $P_l$, we use an RoI Align layer to crop the $P_l$-th feature
map according to bounding box coordinates and thus obtain its features.
Thus, we can extract suitable visual features for multi-scale ground-truth objects.

\paragraph{Improved BBox Evaluation.}
With these multi-scale object features, our model can predict their coordinates of bounding boxes. 
In the object detection scenario, bounding box targets can be formulated as corner-wise coordinates
 $\boldsymbol{b}_i\!=\!(x_1, y_1, x_2, y_2)$ or center-wise coordinates $\boldsymbol{b}^{cen}_i\!\!=\!\!(x_c, y_c, w_c, h_c)$. To avoid coordinate scale issue, 
each coordinate is rescaled to the range [0,1] by using \emph{bounding box center normalization}. Since the bias in the former one (i.e., $x_1 < x_2, y_1 < y_2$) is hard
to be fit by a linear model in LogME, we thus select the latter one. 
LogME for bounding box regression is obtained by averaging over $4$-d coordinates, where each coordinate learns different prior distribution with different $\alpha$ and $\beta$. Considering the
inherent relation between coordinates, we feed $4$-d coordinates of a bounding box as a whole and learn a shared prior distribution for 4 coordinates with the same $\alpha$ and $\beta$. 
More specifically, We expand the dimension of $\boldsymbol{m}$ in Eq. \eqref{equ:logme} from $\boldsymbol{m}\!\!\in\!\!\mathbb{R}^{D}$ to $\boldsymbol{m}\!\!\in\!\!\mathbb{R}^{D\!\times\!4}$ to match the dimension of $\boldsymbol{B}^{cen}$, resulting in a more efficient and accurate evaluation.

\begin{table*}[htbp]
  \centering
  \caption{Ranking results of of six methods for 1\% 33-choose-22 possible source model sets (over 1.9M) on $6$ downstream target datasets. Higher $\tau_w$ and Rel@1 indicate better ranking and transferability metric. As SFDA is specifically designed for classification task, it is not applicable for the single-class task of CrowdHuman. 
  The results of all three variants of our approach, U-LogME, IoU-LogME, and Det-LogME are reported. 
  The best methods are in {\color{red} red} and good ones are in {\color{blue} blue}.}
  \vspace{-3mm}
  \scalebox{0.65}{

\begin{tabular}{l|cccccc|cccccc}
\toprule
Measure & \multicolumn{6}{c}{\emph{Weighted Kendall}'s tau ($\tau_w$) $\uparrow$} & \multicolumn{6}{c}{Top1 Relative Accuracy (Rel@1) $\uparrow$} \\
Method & KNAS  & SFDA  & LogME & U-LogME & IoU-LogME & Det-LogME & KNAS  & SFDA  & LogME & U-LogME & IoU-LogME & Det-LogME \\
\midrule
Pascal VOC & 0.10$\pm$0.18 & \textcolor[rgb]{ 0,  0,  1}{\boldmath{}\textbf{0.65$\pm$0.13}\unboldmath{}} & 0.15$\pm$0.22 & 0.40$\pm$0.17 & 0.58$\pm$0.16 & \textcolor[rgb]{ 1,  0,  0}{\boldmath{}\textbf{0.78$\pm$0.03}\unboldmath{}} & 0.94$\pm$0.10 & \textcolor[rgb]{ 1,  0,  0}{\boldmath{}\textbf{1.00$\pm$0.00}\unboldmath{}} & 0.91$\pm$0.12 & 0.96$\pm$0.05 & \textcolor[rgb]{ 1,  0,  0}{\boldmath{}\textbf{1.00$\pm$0.00}\unboldmath{}} & \textcolor[rgb]{ 1,  0,  0}{\boldmath{}\textbf{1.00$\pm$0.00}\unboldmath{}} \\
CityScapes & -0.22$\pm$0.24 & 0.45$\pm$0.06 & 0.15$\pm$0.20 & 0.13$\pm$0.16 & \textcolor[rgb]{ 0,  0,  1}{\boldmath{}\textbf{0.51$\pm$0.09}\unboldmath{}} & \textcolor[rgb]{ 1,  0,  0}{\boldmath{}\textbf{0.57$\pm$0.08}\unboldmath{}} & 0.95$\pm$0.06 & 0.95$\pm$0.01 & 0.95$\pm$0.06 & 0.90$\pm$0.04 & \textcolor[rgb]{ 1,  0,  0}{\boldmath{}\textbf{0.98$\pm$0.02}\unboldmath{}} & \textcolor[rgb]{ 1,  0,  0}{\boldmath{}\textbf{0.98$\pm$0.02}\unboldmath{}} \\
SODA  & -0.46$\pm$0.09 & 0.46$\pm$0.12 & 0.13$\pm$0.21 & 0.04$\pm$0.17 & \textcolor[rgb]{ 1,  0,  0}{\boldmath{}\textbf{0.61$\pm$0.09}\unboldmath{}} & \textcolor[rgb]{ 1,  0,  0}{\boldmath{}\textbf{0.61$\pm$0.09}\unboldmath{}} & 0.88$\pm$0.04 & 0.95$\pm$0.02 & 0.92$\pm$0.11 & 0.87$\pm$0.06 & \textcolor[rgb]{ 1,  0,  0}{\boldmath{}\textbf{0.98$\pm$0.02}\unboldmath{}} & \textcolor[rgb]{ 1,  0,  0}{\boldmath{}\textbf{0.98$\pm$0.02}\unboldmath{}} \\
CrowdHuman & -0.42$\pm$0.11 & N/A   & 0.19$\pm$0.19 & 0.21$\pm$0.18 & \textcolor[rgb]{ 1,  0,  0}{\boldmath{}\textbf{0.34$\pm$0.16}\unboldmath{}} & \textcolor[rgb]{ 1,  0,  0}{\boldmath{}\textbf{0.34$\pm$0.16}\unboldmath{}} & 0.85$\pm$0.04 & N/A   & 0.97$\pm$0.04 & 0.97$\pm$0.04 & \textcolor[rgb]{ 1,  0,  0}{\boldmath{}\textbf{0.98$\pm$0.03}\unboldmath{}} & \textcolor[rgb]{ 1,  0,  0}{\boldmath{}\textbf{0.98$\pm$0.03}\unboldmath{}} \\
VisDrone & 0.04$\pm$0.20 & 0.53$\pm$0.12 & 0.48$\pm$0.17 & 0.17$\pm$0.17 & \textcolor[rgb]{ 1,  0,  0}{\boldmath{}\textbf{0.70$\pm$0.08}\unboldmath{}} & \textcolor[rgb]{ 0,  0,  1}{\boldmath{}\textbf{0.69$\pm$0.08}\unboldmath{}} & 0.88$\pm$0.17 & \textcolor[rgb]{ 1,  0,  0}{\boldmath{}\textbf{1.00$\pm$0.00}\unboldmath{}} & 0.90$\pm$0.15 & 0.78$\pm$0.13 & \textcolor[rgb]{ 0,  0,  1}{\boldmath{}\textbf{0.99$\pm$0.02}\unboldmath{}} & \textcolor[rgb]{ 0,  0,  1}{\boldmath{}\textbf{0.99$\pm$0.02}\unboldmath{}} \\
DeepLesion & -0.13$\pm$0.19 & -0.21$\pm$0.14 & 0.08$\pm$0.20 & \textcolor[rgb]{ 1,  0,  0}{\boldmath{}\textbf{0.52$\pm$0.14}\unboldmath{}} & -0.05$\pm$0.18 & \textcolor[rgb]{ 0,  0,  1}{\boldmath{}\textbf{0.42$\pm$0.17}\unboldmath{}} & 0.69$\pm$0.11 & 0.65$\pm$0.06 & 0.72$\pm$0.17 & \textcolor[rgb]{ 1,  0,  0}{\boldmath{}\textbf{0.87$\pm$0.28}\unboldmath{}} & 0.64$\pm$0.07 & \textcolor[rgb]{ 0,  0,  1}{\boldmath{}\textbf{0.75$\pm$0.38}\unboldmath{}} \\
\midrule
Average & -0.18$\pm$0.28 & 0.31$\pm$0.10 & 0.19$\pm$0.20 & 0.24$\pm$0.17 & \textcolor[rgb]{ 0,  0,  1}{\boldmath{}\textbf{0.45$\pm$0.13}\unboldmath{}} & \textcolor[rgb]{ 1,  0,  0}{\boldmath{}\textbf{0.57$\pm$0.10}\unboldmath{}} & 0.86$\pm$0.09 & 0.91$\pm$0.02 & 0.90$\pm$0.11 & 0.89$\pm$0.10 & \textcolor[rgb]{ 0,  0,  1}{\boldmath{}\textbf{0.93$\pm$0.03}\unboldmath{}} & \textcolor[rgb]{ 1,  0,  0}{\boldmath{}\textbf{0.95$\pm$0.08}\unboldmath{}} \\
\bottomrule
\end{tabular}%

}
  \label{tab:main_results}%
\end{table*}%

\paragraph{Unified Sub-task Evaluation.}
Although the correlations among the coordinates are captured by joint evaluation of bounding box coordinates, LogME for regression sub-task still suffers from neglecting object classes information.
So how to build the correlation between these sub-tasks?
Inspired by the class-aware detection heads of classical detectors \cite{ren2015faster, tian2019fcos, zhu2020deformable}, we propose \emph{unified sub-task evaluation} for assessing the transferability of pre-trained detectors.
To be specific, we combine the bounding box matrix $\boldsymbol{B}^{cen}$ and class label matrix $\boldsymbol{C}$ as the unified label matrix $\boldsymbol{Y}^u$.
The bounding box matrix $\boldsymbol{B}^{cen}$ is expanded from $\boldsymbol{B}^{cen}\!\!\in\!\!\mathbb{R}^{M\!\times\!4}$ to $\boldsymbol{Y}^u\!\!\in\!\!\mathbb{R}^{M\!\times\!(4 \cdot K)}$ according to $\boldsymbol{C}$, where $K$ is the total number of classes.
By integrating pseudo bounding boxes filled with $0$ coordinates, we obtain unified label matrix $\boldsymbol{Y}^u$ as
\vspace{-2mm}
\begin{equation}
\centering
\small
\boldsymbol{Y}^u\!\!=\!\!\left[\!\left[\!\underbrace{(0,0,0,0)}_{1\operatorname{st}},\dots\!,\overbrace{\underbrace{\left(x_c, y_c, w_c, h_c\right)}_{c_{i}\operatorname{-th}}}^{\boldsymbol{b}_{i}^{cen}},\dots\!,\underbrace{(0,0,0,0)}_{K\operatorname{-th}}\!\right]\!\right]_M\!\!\!.
    \label{equ:unified_y}
\end{equation}
To this end, both bounding boxes and classes information are represented by $\boldsymbol{Y}^u$.
Accordingly, $\boldsymbol{m}$ in Eq. \eqref{equ:logme} is further expanded from $\boldsymbol{m}\!\!\in\!\!\mathbb{R}^{D\!\times\!4}$ to $\boldsymbol{m}\!\!\in\!\!\mathbb{R}^{D\!\times\!(4 \cdot K)}$ for matching the dimension of $\boldsymbol{Y}^{u}$.
We can take the unified label matrix $\boldsymbol{Y}^u$ as the input for transferability assessment and obtain a unified transferability score for detection as the following:
\begin{equation}
\small
    \operatorname{U-LogME} = \log p(\boldsymbol{Y}^u\!\mid\!\boldsymbol{F}).
    \label{equ:u_logme}
\end{equation}


\subsection{IoU-LogME}
\label{sec:iou_logme}

Although U-LogME addressed multi-scale and multi-task issues of vanilla LogME, the MSE used for fitting features and bounding box coordinates is not robust to object scales. That is, the larger objects might cause bigger MSE, which makes the model prefers larger objects than smaller ones. Moreover, each coordinate is evaluated separately in mean squared error, without considering the correlations between different coordinates. To overcome this issue, we introduce IoU metric into U-LogME. Specifically, $\boldsymbol{m}$ in Eq. \eqref{equ:logme} is interpreted as a linear regression model so that $\boldsymbol{F}\boldsymbol{m}$ can be regarded as the bounding box predictions from a detector naturally.
We propose to calculate the IoU between the bounding boxes predictions $\boldsymbol{F}\boldsymbol{m}$ and the corresponding ground truth bounding boxes $\boldsymbol{B}^{cen}$ as a IoU based transferability measurement.
Considering $\boldsymbol{m}\!\!\in\!\!\mathbb{R}^{D\times (4\cdot K)}$ is computed from the unified label matrix $\boldsymbol{Y}^u\!\!\in\!\!\mathbb{R}^{M\times (4 \cdot K)}$ in Eq. \eqref{equ:unified_y}, so we downsample $\boldsymbol{m}$ to $\boldsymbol{m}'\!\!\in\!\!\mathbb{R}^{D\times 4}$ by reserving the values where real coordinates of $\boldsymbol{B}^{cen}$ arise.
This IoU-based metric is formulated as:
\begin{equation}
\small
    \operatorname{IoU-LogME} = \operatorname{IoU}(\boldsymbol{F}\boldsymbol{m}', \boldsymbol{B}^{cen}) = \frac{|\boldsymbol{F}\boldsymbol{m}' \cap \boldsymbol{B}^{cen}|}{|\boldsymbol{F}\boldsymbol{m}' \cup \boldsymbol{B}^{cen}|}.
    \label{equ:iou_logme}
\end{equation}

\subsection{Det-LogME}

Although IoU-LogME is invariant to objects with different scales, it degrades to 0 when the two inputs have no intersection and fails to measure the absolute difference between them. On the contrary, MSE is good at tackling these cases. Therefore, to take advantage of their strengths, we propose to combine them together and obtain a final detector assessment metric $\operatorname{Det-LogME}$, which is formulated as:
\begin{equation}
\small
\operatorname{Det-LogME} = \operatorname{U-LogME} + \mu \cdot \operatorname{IoU-LogME},
    \label{equ:det_logme}
\end{equation}
where $\mu$ is used for controlling the weight of IoU.
$\operatorname{U-LogME}$ and $\operatorname{IoU-LogME}$ are normalized to $[0,1]$ upon 33 pre-trained detectors to unify the scale.

\section{Experiment}

\subsection{Experimental Setup}
\label{sec:experimental_setup}

We employ the proposed benchmark in Sec. \ref{sec:benchmark} to conduct experiments. Our method is compared with KNAS \cite{xu2021knas}, SFDA \cite{shao2022not}, and LogME \cite{you2021logme}. KNAS is a gradient-based approach that operates under the assumption that gradients can predict downstream training performance. Therefore, we use it as a point of comparison with our efficient gradient-free approach. SFDA is not applicable for the single-class task of CrowdHuman \cite{shao2018crowdhuman}. Therefore, we present the SFDA results on five other multiple-class datasets.
It is worth noting that our proposed pyramid feature matching enhances both LogME and SFDA, facilitating their evaluation. Furthermore, we present results for all three variants of our approach: U-LogME, IoU-LogME, and Det-LogME. We utilize two evaluation measures, namely the ranking correlation \emph{Weighted Kendall}'s tau ($\tau_w$), as defined in Eq. \eqref{equ:tw}, and the Top-1 Relative Accuracy (Rel@1). Larger values of $\tau_w$ and Rel@$1$ indicate better assessment results.


\subsection{Main Results}
\label{sec:eval_supervised}

As discussed in \cite{agostinelli2022stable}, transferability metrics can be unstable. To address this issue, we randomly sub-sample 22 models from 33 pre-trained detectors and use 1\% of the 33-choose-22 possible source model sets (over 1.9M) for measurement, as shown in Table \ref{tab:main_results}.
Our observations indicate that KNAS performs poorly on all 6 target datasets, with negative ranking correlation $\tau_w$. Our method Det-LogME consistently outperforms LogME on 6 downstream tasks. For instance, Det-LogME surpasses LogME by substantial margins of $0.63$, $0.48$, and $0.34$ in terms of ranking correlation $\tau_w$ on Pascal VOC, SODA, and DeepLesion, respectively. This validates the high effectiveness of our proposed unified evaluation and complementary IoU measurement.
On five multiple-class downstream datasets, Det-LogME still outperforms classification-specific SFDA, particularly on DeepLesion ($+0.63 \ \tau_w$). This further indicates that classification-specific transferability metrics may not yield optimal results for multi-class detection problems.
Furthermore, our Det-LogME achieves better Rel@1 values than all other SOTA methods in average, outperforming the second-best method SFDA by 4\% top-1 relative accuracy. Overall, our method is robust and consistently outperforms competing methods over 1.9M sub-sampled model sets.

\begin{figure}[t]
  \centering
   \includegraphics[width=\linewidth]{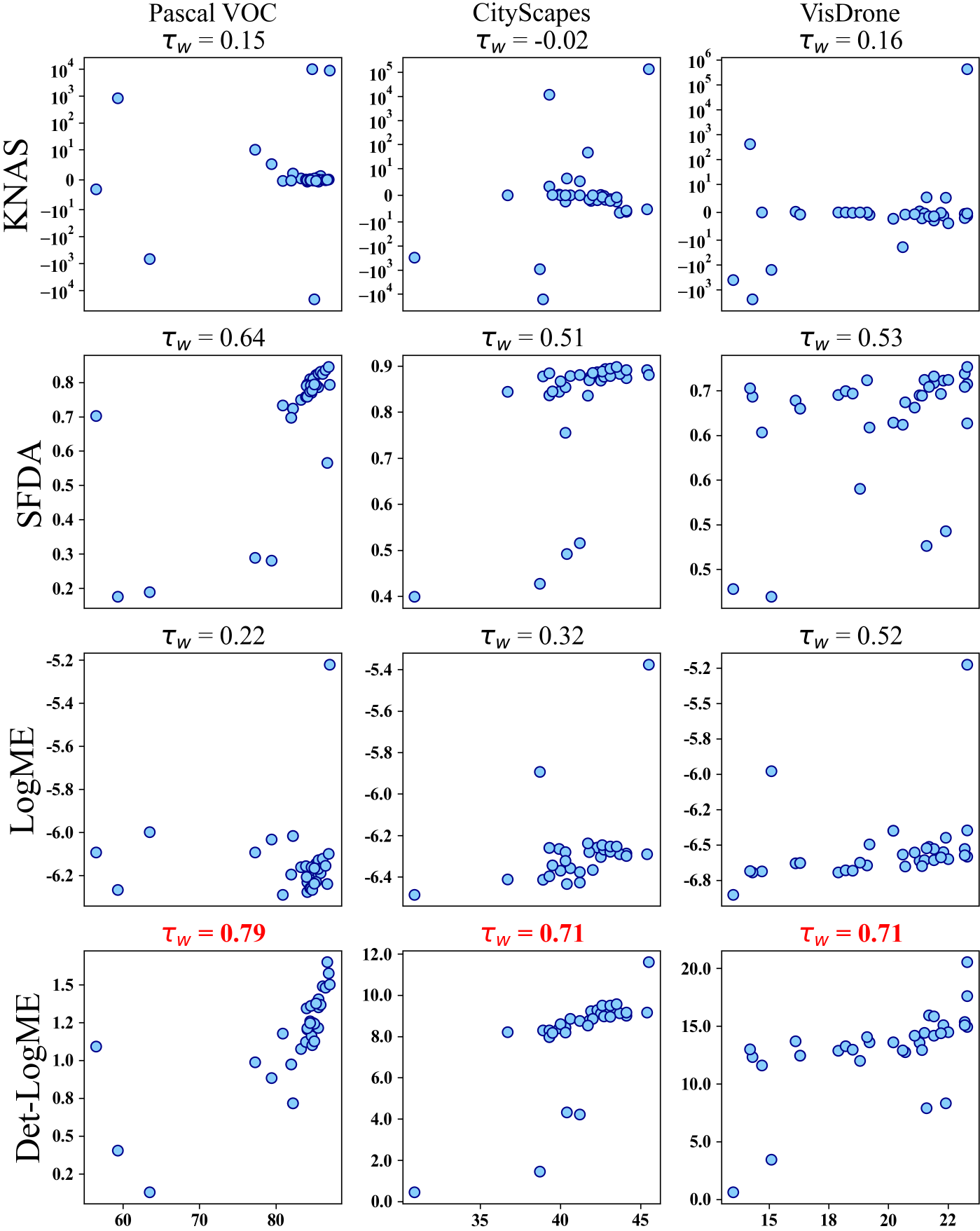}
   \vspace{-7mm}
   \caption{
    Comparison of ranking scores.
    The plots illustrate ground-truth fine-tuning performance $\{g_n\}_{n=1}^N$ (x-axis), ranking scores (y-axis), and \emph{Weighted Kendall}’s coefficient $\tau_w$ for 33 pre-trained detectors on 3 out of 6 target datasets.}
   \label{fig:ranking_tw_all}
\end{figure}

Regarding the three variants proposed in this paper, U-LogME performs the worst in most cases. We have further observations to make.
On one hand, IoU-LogME, which uses IoU as a scale-invariant metric, performs better than MSE-based U-LogME in scenarios where the objects vary greatly in scale. This is especially evident in CityScapes \cite{cordts2016cityscapes}, SODA \cite{han2021soda10m}, and VisDrone \cite{zhu2021detection}, where objects captured under driving or UAV scenarios have diverse scales.
On the other hand, IoU-based IoU-LogME suffers from the problem of remaining equal to $0$ when there is no intersection between the predicted and ground truth bounding boxes, regardless of how far apart they are. In contrast, MSE-based U-LogME can handle this problem with absolute distances. This is particularly evident in DeepLesion \cite{yan2018deeplesion}, where the lesions are very small and difficult to detect.
Det-LogME, which combines the advantages of both unified evaluation and IoU measurement, achieves a trade-off and better ranking performance than U-LogME and IoU-LogME. We can also observe that Det-LogME performs better on five multiple-class tasks than the single-class one. This indicates the significant difficulties and challenges involved in evaluating the pre-trained detectors on dense tasks.

Figure \ref{fig:ranking_tw_all} illustrates the relationship between predicted transferability and actual fine-tuning performance of 33 pre-trained detectors for our Det-LogME and three SOTA methods on three target datasets. We observe that our Det-LogME consistently shows better positive correlations compared to other SOTA methods in all experiments, which demonstrates the effectiveness of our proposed approach.

\subsection{Ablation Studies}
In this subsection, we carefully study our proposed Det-LogME with respect to different components and hyper-parameters by assessing $33$ detectors on the Pascal VOC.

\paragraph{Components Analysis of Det-LogME.}
As shown in Table \ref{tab:det_logme_components}, we can learn that the large variances of different coordinates are eliminated by bounding box center normalization and the ranking correlation $\tau_w$ is improved $0.11$.
By jointly evaluating $4$-d coordinates, $\tau_w$ improves from $0.33$ to $0.40$, indicating the inherent correlations among $4$ coordinates within a bounding box are captured by our proposed joint evaluation.
Under unified sub-task evaluation, $\tau_w$ further improves from $0.40$ to $0.43$, which demonstrates the effectiveness of unifying the supervision information by considering object classes information.
Finally, we observe that the inclusion of the complementary IoU metric in our evaluation framework leads to a substantial increase in ranking performance from 0.43 to 0.79 ($+0.36\ \tau_w$). This finding underscores the importance of IoU measurement in assessing the transferability of pre-trained detectors.

\begin{table}[t]
\centering
\begin{minipage}{0.48\linewidth}
\caption{Effects of different components of Det-LogME.}
\vspace{-3mm}
\label{tab:det_logme_components}
		\centering
		\makeatletter\def\@captype{table}\makeatother
  \scalebox{0.85}{	
\begin{tabular}{lc}
\toprule
Method & $\tau_w$ $\uparrow$ \\
\midrule
Baseline (LogME) & 0.22  \\
\midrule
w/ bbox center norm. & 0.33  \\
w/ joint eval. of coord. & 0.40  \\
w/ unified sub-task eval. & 0.43  \\
w/ IoU (Det-LogME) & \textbf{0.79} \\
\bottomrule
\end{tabular}%
}
	
 \end{minipage}\quad
\begin{minipage}{0.47\linewidth}
\caption{Effects of different bounding box normalization techniques (border and center) in Det-LogME.}
\vspace{-3mm}
\label{tab:bbox_normalization}
		\centering
		\makeatletter\def\@captype{table}\makeatother
		
	\scalebox{0.85}{	
		\begin{tabular}{lc}
\toprule
Method & $\tau_w$ $\uparrow$ \\
\midrule
Baseline (LogME) & 0.22  \\
\midrule
w/ bbox border norm. & 0.22  \\
w/ bbox center norm. & \textbf{0.33} \\
\bottomrule
\end{tabular}%
}

\end{minipage}
\end{table}

\begin{table*}[t]
\caption{Efficiency evaluation of Det-LogME and comparison with brute-force fine-tuning, naive feature extraction, KNAS, SFDA, and LogME. Note that the wall-clock time and memory footprint for SFDA are evaluated on $5$ multiple-class downstream detection datasets.}
\tabcolsep=1.2mm
\vspace{-3mm}
\label{tab:efficiency}%
  \centering
  \scalebox{0.75}{

\begin{tabular}{l|cc|cccc|cc|cccc}
\toprule
Measure & \multicolumn{6}{c|}{Wall-clock Time (s) $\downarrow$}      & \multicolumn{6}{c}{Memory Footprint (GB) $\downarrow$} \\
\midrule
\multirow{2}[2]{*}{Method} & Fine-tune & Extract feature & \multirow{2}[2]{*}{KNAS} & \multirow{2}[2]{*}{SFDA} & \multirow{2}[2]{*}{LogME} & \multirow{2}[2]{*}{Det-LogME} & Fine-tune & Extract feature & \multirow{2}[2]{*}{KNAS} & \multirow{2}[2]{*}{SFDA} & \multirow{2}[2]{*}{LogME} & \multirow{2}[2]{*}{Det-LogME} \\
      &  (upper bound) &  (lower bound) &       &       &       &       &  (upper bound) &  (lower bound) &       &       &       &  \\
\midrule
Pascal VOC & 28421.09  & 128.88  & 129.99  & 130.14  & 130.18  & 129.89  & 13.39  & 0.47  & 13.86  & 0.55  & 0.59  & 0.66  \\
CityScapes & 73973.09  & 50.50  & 51.53  & 52.02  & 51.80  & 51.18  & 22.36  & 0.99  & 23.34  & 1.06  & 1.08  & 1.10  \\
SODA  & 13928.24  & 50.75  & 51.89  & 51.88  & 51.83  & 51.19  & 12.54  & 0.50  & 13.04  & 0.54  & 0.55  & 0.55  \\
CrowdHuman & 43297.45  & 170.38  & 171.48  & N/A   & 178.26  & 175.06  & 27.24  & 0.63  & 27.87  & N/A   & 1.47  & 1.46  \\
VisDrone & 14067.15  & 63.38  & 64.51  & 69.08  & 68.16  & 64.97  & 16.69  & 0.65  & 17.34  & 0.79  & 0.80  & 0.84  \\
DeepLesion & 8465.45  & 37.13  & 38.24  & 37.49  & 37.71  & 37.34  & 6.99  & 0.52  & 7.51  & 0.53  & 0.54  & 0.54  \\
\midrule
Average & 30358.75  & 83.50  & 84.60  & 68.12  & 86.32  & 84.94  & 16.53  & 0.63  & 17.16  & 0.69  & 0.84  & 0.86  \\
\bottomrule
\end{tabular}%

}

\end{table*}%

\begin{figure}[t]
  \centering
   \includegraphics[width=0.75\linewidth]{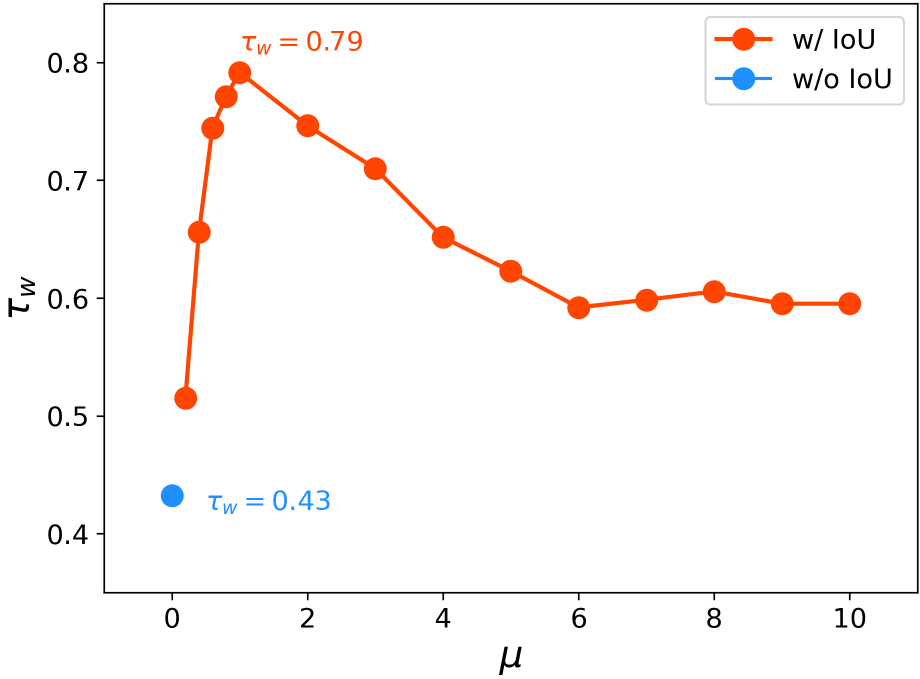}
   \vspace{-4mm}
   \caption{Effects of different weights $\mu$ for complementary IoU metric in Det-LogME. The first blue marker indicates Det-LogME without IoU measurement (degrades to U-LogME).}
   \label{fig:iou_weight}
\end{figure}

\paragraph{Different BBox Normalization Techniques.}
To mitigate the detrimental effects of large variances among different coordinates, we introduce bounding box normalization. We experiment with two widely used normalization techniques in detection. Except for center normalization described in Section \ref{sec:u_logme}, we also try to apply border normalization by dividing the bounding box with the corresponding width and height of the input image to obtain $\boldsymbol{b}^{bor}_i\!\!=\!\!(x_1', y_1', x_2', y_2')$. From the results shown in Table \ref{tab:bbox_normalization}, we observe that border normalization yields no improvement, and center normalization outperforms it by $+0.1 \ \tau_w$. Center normalization scales the four coordinates of the bounding box to the same range in $[0, 1]$, while border normalization may not work well when significant biases exist among the $x$-axis and $y$-axis coordinates, such as when an object is located near the top-right corner of an image where $x_1 \gg y_1$ or $x_2 \gg y_2$.

  






\paragraph{Weight of Complementary IoU Metric.}
The weight hyper-parameter $\mu$ in Eq. \eqref{equ:det_logme} controls the behavior of the complementary IoU metric. Similar to the weights of IoU-related losses \cite{rezatofighi2019generalized, zheng2020distance}, the weight $\mu$ is crucial to the final transferability score of Det-LogME. We investigate the effects of different weights $\mu\!\!\in\!\!\{0, 0.2, \dots, 0.8, 1, 2, \dots, 10\}$ of the IoU metric, and the results are presented in Figure \ref{fig:iou_weight}. We observe that incorporating the IoU metric as a complementary term consistently improves the ranking performance of Det-LogME compared to not including it (the first blue marker, which degrades to U-LogME, with $\tau_w=0.43$). The best ranking performance of Det-LogME is achieved when the weight is $1$, resulting in $\tau_w=0.79$.

\subsection{Efficiency Analysis}
\label{sec:efficiency}
In this subsection, we present a comprehensive evaluation of the assessing efficiency of Det-LogME and compare it with brute-force fine-tuning, naive feature extraction, KNAS, SFDA, and LogME from two perspectives:
1) wall-clock time: the average time of fine-tuning (12 epochs) or evaluating (including feature extraction) all $33$ pre-trained detectors;
2) memory footprint: the maximum memory required during fine-tuning or evaluating (including feature extraction, the loading of all visual features, and computing transferability metric) all $33$ pre-trained detectors.
The efficiency of classification-specific SFDA is studied on $5$ multiple-class tasks.
The results are presented in Table \ref{tab:efficiency}.

\paragraph{Wall-clock Time.} 
Over 6 downstream tasks, KNAS has the fastest speed, but according to Table \ref{tab:main_results}, it performs the worst, which is not acceptable in practice. On the other hand, with joint evaluation of bounding box coordinates, Det-LogME runs faster than LogME because Det-LogME does not need to evaluate all $4$-d coordinates in a loop. Moreover, Det-LogME runs faster than SFDA, which includes fine-tuning dynamics on all $5$ multiple-class datasets. Considering that the selected model will be fine-tuned on the target task, our proposed Det-LogME brings about $(\operatorname{NumSourceModels} -1) \times$ speedup compared with brute-force fine-tuning, which is $32\times$ in this work.

\paragraph{Memory Footprint.} 
Table \ref{tab:efficiency} indicates that gradient-based KNAS demands a significant amount of memory exceeding 17 GB, rendering it infeasible for most practical applications. In comparison, Det-LogME only marginally increases the memory when compared to LogME and SFDA, making it practical for deployment. Furthermore, Det-LogME demonstrates high memory efficiency by requiring 19$\times$ less memory than brute-force fine-tuning, underscoring its potential to save computational resources.


\section{Conclusion}
In this paper, we aim to address a practical but underexplored problem of efficient transferability assessment for pre-trained detection models. 
To achieve this, we establish a challenging detector transferability benchmark comprising a large and diverse zoo consisting of 33 detectors with various architectures, source datasets and schemes. 
Upon this zoo, we adopt 6 downstream tasks spanning 5 diverse domains for evaluation. 
Further, we propose a simple yet effective framework for assessing the transferability of pre-trained detectors. 
Extensive experimental results demonstrate the high effectiveness and efficiency of our approach compared with other state-of-the-art methods across a wide range of pre-trained detectors and downstream tasks, notably outperforming brute-force fine-tuning in terms of computational efficiency. 
We hope our work can inspire further research into the selection of pre-trained models, particularly those with multi-scale and multi-task capabilities.

\medskip
\paragraph{\bf Acknowledgements.} We gratefully acknowledge the support of Mindspore, CANN (Compute Architecture for Neural Networks) and Ascend AI Processor used for this work.

\appendix


\section{LogME Measurement for Object Detection}
\label{sec:appendix_logme}

In this work, we extend a classification assessment method LogME \cite{you2021logme} to object detection. In this section, we will give detailed derivations of LogME for object detection framework.

Different from image-level features used for assessing classification task, we extract object-level features of ground-truth bounding boxes by using pre-trained detectors' backbone followed by an ROIAlign layer \cite{ren2015faster}. In this way, for a given pre-trained detector and a downstream task, we can collect the object-level features of downstream task by using the detector and form a feature matrix $\boldsymbol{F}$, with each row $\boldsymbol{f}_i$ denotes an object-level feature vector. For each $\boldsymbol{f}_i$, we also collect its 4-d coordinates of grounding-truth bounding box $\boldsymbol{b}_i$ and class label $c_i$ to form a bounding box matrix $\boldsymbol{B}$ and a class label matrix $\boldsymbol{C}$.

For the bounding box regression sub-task, LogME measures the transferability by using the maximum evidence $p(\boldsymbol{B}|\boldsymbol{F})\!\!=\!\!\int p(\boldsymbol{\theta}|\alpha)p(\boldsymbol{B}|\boldsymbol{F},\beta,\boldsymbol{\theta}) d\boldsymbol{\theta}$, where $\boldsymbol{\theta}$ is the parameter of linear model. $\alpha$ denotes the parameter of prior distribution of $\boldsymbol{\theta}$, and $\beta$ denotes the parameter of posterior distribution of each observation $p(\boldsymbol{b}_i|\boldsymbol{f}_i,\beta,\boldsymbol{\theta})$. 
By using the evidence theory \cite{knuth2015bayesian} and basic principles in graphical models \cite{koller2009probabilistic}, the evidence can be calculated as
\begin{equation}
\small
\begin{aligned}
p(\boldsymbol{B}\!\mid\!\boldsymbol{F}) &=\int p(\boldsymbol{\theta}|\alpha)p(\boldsymbol{B}|\boldsymbol{F},\beta,\boldsymbol{\theta}) d\boldsymbol{\theta} \\
&=\int p(\boldsymbol{\theta}\!\mid\!\alpha) \prod^M_{i=1} p(\boldsymbol{b}_i|\boldsymbol{f}_i,\beta,\boldsymbol{\theta}) d\boldsymbol{\theta}\\
&=\left(\frac{\beta}{2 \pi}\right)^{\frac{M}{2}}\left(\frac{\alpha}{2 \pi}\right)^{\frac{D}{2}} \int e^{-\frac{\alpha}{2} \boldsymbol{\theta}^T \boldsymbol{\theta}-\frac{\beta}{2}\|\boldsymbol{f}_i\boldsymbol{\theta}-\boldsymbol{b}_i\|^2} \mathrm{~d} \boldsymbol{\theta},
\end{aligned}
\label{equ:appendix_evidence}
\end{equation}
where $M$ is the number of objects and $D$ is the dimension of object features.
When $A$ is positive definite, 
\begin{equation}
\small
\int e^{-\frac{1}{2}\left(\boldsymbol{\theta}^T A \boldsymbol{\theta}+b^T \boldsymbol{\theta}+c\right)} \mathrm{d} \boldsymbol{\theta}=\frac{1}{2}\sqrt{\frac{(2 \pi)^D}{|A|}} e^{\frac{1}{4} b^T A^{-1} b- c}.
\end{equation}

LogME takes the logarithm of Eq. \eqref{equ:appendix_evidence} for simpler calculation. So the transferability score is expressed by 
\begin{equation}
\small
\begin{aligned}
    \text{LogME} =
    &\log p(\boldsymbol{B}|\boldsymbol{F}) \\
    =&\frac{M}{2} \log \beta+\frac{D}{2} \log \alpha-\frac{M}{2} \log 2 \pi \\
    &-\frac{\beta}{2}\|\boldsymbol{F} \boldsymbol{m}-\boldsymbol{B}\|_2^2-\frac{\alpha}{2} \boldsymbol{m}^T \boldsymbol{m}-\frac{1}{2} \log |A|.
\end{aligned}
\label{equ:appendix_logme}
\end{equation}
where $A$ and $m$ are
\begin{equation}
\small
A=\alpha I+\beta \boldsymbol{F}^T \boldsymbol{F}, \boldsymbol{m}=\beta A^{-1} \boldsymbol{F}^T\boldsymbol{B},
\label{equ:appendix_Am}
\end{equation}
where $A$ is the $L_2$-norm of $\boldsymbol{F}$, and $\boldsymbol{m}$ is the solution of $\boldsymbol{\theta}$.
Here $\alpha$ and $\beta$ are maxmized by alternating between evaluating $\boldsymbol{m}$, $\gamma$ and maximizing $\alpha$, $\beta$ with $\boldsymbol{m}$, $\gamma$ fixed \cite{Gull1989} as the following:
\begin{equation}
\small
\begin{gathered}
    \gamma=\sum_{i=1}^D \frac{\beta \sigma_i}{\alpha+\beta \sigma_i},
    \alpha \leftarrow \frac{\gamma}{\boldsymbol{m}^T \boldsymbol{m}}, 
    \beta \leftarrow \frac{M-\gamma}{\|\boldsymbol{F} \boldsymbol{m}-\boldsymbol{B}\|_2^2},
\end{gathered}
\label{equ:appendix_optim}
\end{equation}
where $\sigma_i$'s are singular values of $\boldsymbol{F}^T \boldsymbol{F}$.
With the optimal $\alpha^*$ and $\beta^*$, the  logarithm maximum evidence $\mathcal{L}\left(\alpha^*, \beta^*\right)$ is used for evaluating the transferability.
Considering $\mathcal{L}\left(\alpha^*, \beta^*\right)$ scales linearly with the number of objects $M$, it is normalized as $\frac{\mathcal{L}\left(\alpha^*, \beta^*\right)}{M}$, which is interpreted as the average logarithm maximum evidence of all given object feature matrix $\boldsymbol{F}$ and bounding box matrix $\boldsymbol{B}$.
LogME for classification sub-task can be computed by replacing $\boldsymbol{B}$ in Eq. \eqref{equ:appendix_logme} with converted one-hot class label matrix. 

Nevertheless, optimizing LogME by Eq. \eqref{equ:appendix_Am} and Eq. \eqref{equ:appendix_optim} is timely costly, which is comparable with brute-force fine-tuning.
So LogME further improves the computation efficiency as follows.
The most expensive steps in Eq. \eqref{equ:appendix_Am} are to calculate the inverse matrix $A^{-1}$ and matrix multiplication $A^{-1} \boldsymbol{F}^T$, which can be avoided by decomposing $\boldsymbol{F}^T \boldsymbol{F}$.
The decomposition is taken by $\boldsymbol{F}^T\boldsymbol{F}=V \operatorname{diag}\{\sigma\} V^T$, where $V$ is an orthogonal matrix.
By taking $\Lambda=\operatorname{diag}\{(\alpha+\beta \sigma)\}$, $A$ and $A^{-1}$ turn to $A=\alpha I+\beta \boldsymbol{F}^T\boldsymbol{F}=V \Lambda V^T$ and $A^{-1}=V \Lambda^{-1} V^T$.
With associate law, LogME takes a fast computation by $A^{-1}\boldsymbol{F}^T\boldsymbol{B}=\left(V\left(\Lambda^{-1}\left(V^T\left(\boldsymbol{F}^T\boldsymbol{B}\right)\right)\right)\right)$.
To this end, the computation of $\boldsymbol{m}$ in Eq. \eqref{equ:appendix_Am} is optimized as 
\begin{equation}
\small
\boldsymbol{m}=\beta\left(V\left(\Lambda^{-1}\left(V^T\left(\boldsymbol{F}^T \boldsymbol{B}\right)\right)\right)\right).
\label{equ:appendix_optimized_m}
\end{equation}

\begin{algorithm}[t]
\small
\caption{Det-LogME}
\label{alg:det_logme}
\textbf{Input:} pre-trained detector $\mathcal{F}$, target dataset $\mathcal{D}_t$\\
\textbf{Output:} estimated transferability score $\operatorname{Det-LogME}$
\begin{algorithmic}[1] 
\State Extract multi-scale object-level features using pre-trained detector $\mathcal{F}$’s backbone followed by an ROIAlign layer and collect bounding box coordinates and class labels:

$\boldsymbol{F}\!\in\!\mathbb{R}^{M \times D}$, $\boldsymbol{B}\!\in\!\mathbb{R}^{M \times 4}$, $\boldsymbol{C}\!\in\!\mathbb{R}^{M}$
\State Find the match level features for all objects
\State Apply center normalization on $\boldsymbol{B}$ to obtain $\boldsymbol{B}^{cen}$
\State Unify $\boldsymbol{B}^{cen}$ and $\boldsymbol{C}$ as a unified label matrix $\boldsymbol{Y}^u$ by

\noindent$\boldsymbol{Y}^u\!\!=\!\!\left[\!\left[\!\underbrace{(0,0,0,0)}_{1\operatorname{st}},\dots\!,\overbrace{\underbrace{\left(x_c, y_c, w_c, h_c\right)}_{c_{i}\operatorname{-th}}}^{\boldsymbol{b}_{i}^{cen}},\dots\!,\underbrace{(0,0,0,0)}_{K\operatorname{-th}}\!\right]\!\right]_M\!\!\!$
\State Initialize $\alpha=1, \beta=1$, compute $\boldsymbol{F}^T \boldsymbol{F}=V \operatorname{diag}\{\sigma\} V^T$ 

\While{$\alpha$ and $\beta$ not converge}
\State Compute $\gamma=\sum_{i=1}^D \frac{\beta \sigma_i}{\alpha+\beta \sigma_i}$, $\Lambda=\operatorname{diag}\{(\alpha+\beta \sigma)\}$
\State Compute $\boldsymbol{m}=\beta\left(V\left(\Lambda^{-1}\left(V^T\left(\boldsymbol{F}^T \boldsymbol{B}^{cen}\right)\right)\right)\right)$
\State Expand $\boldsymbol{m}\!\in\!\mathbb{R}^{D}$ to $\boldsymbol{m}\!\in\!\mathbb{R}^{D\!\times\!(4 \cdot K)}$ for matching $\boldsymbol{Y}^{u}$
\State Update $\alpha \leftarrow \frac{\gamma}{\boldsymbol{m}^T \boldsymbol{m}}$, 
$\beta \leftarrow \frac{M-\gamma}{\|\boldsymbol{F} \boldsymbol{m}-\boldsymbol{B}^{cen}\|_2^2}$
\EndWhile

\State Compute $\operatorname{U-LogME}$ by
\vspace{-2mm}
\begin{equation*}
\small
\begin{aligned}
    \operatorname{U-LogME} 
    =&\frac{M}{2} \log \beta+\frac{D}{2} \log \alpha-\frac{M}{2} \log 2 \pi \\
    &-\frac{\beta}{2}\|\boldsymbol{F} \boldsymbol{m}-\boldsymbol{B}^{cen}\|_2^2-\frac{\alpha}{2} \boldsymbol{m}^T \boldsymbol{m}-\frac{1}{2} \log |A|,
\end{aligned}
\vspace{-2mm}
\end{equation*}
where $A=\alpha I+\beta \boldsymbol{F}^T \boldsymbol{F}$

\State Downsample $\boldsymbol{m}$ to $\boldsymbol{m}'\in\mathbb{R}^{D \times 4}$ by reserving the real coordinates of $\boldsymbol{B}^{cen}$, compute $\operatorname{IoU-LogME}=\frac{\left|\boldsymbol{F} \boldsymbol{m}^{\prime} \cap \boldsymbol{B}^{c e n}\right|}{\left|\boldsymbol{F} \boldsymbol{m}^{\prime} \cup \boldsymbol{B}^{c e n}\right|}$

\State Compute $\operatorname{Det-LogME}=\operatorname{U-LogME}+\mu\cdot\operatorname{IoU-LogME}$

\State {\bf Return} $\operatorname{Det-LogME}$

\end{algorithmic}
\end{algorithm}

\section{Details of Experiment Setup}
\label{sec:appendix_moredetails}

In this section, we include more details of our experiment setup, including the source models and target datasets.

\paragraph{Implementation Details.}
Our implementation is based on MMDetection \cite{chen2019mmdetection} with PyTorch 1.8 \cite{paszke2019pytorch} and all experiments are conducted on $8$ V100 GPUs.
The base feature level $l_0$ in \emph{Pyramid Feature Matching}
is set as $3$.
The ground truth ranking of these detectors are obtained by fine-tuning all of them on the downstream tasks with well tuned training hyper-parameters.
The overall Det-LogME algorithm is given in Algorithm \ref{alg:det_logme}.




\paragraph{Baseline Methods.}
We adopt $3$ SOTA methods, KNAS \cite{xu2021knas}, SFDA \cite{shao2022not}, and LogME \cite{you2021logme}, as the baseline methods and make comparisons with our proposed method.
KNAS is a gradient based method different from recent efficient assessment method, we take it as a comparison with our gradient free approach.
SFDA is the current SOTA method on the classification task, so we formulate the multi-class object detection as a object-level classification task for adapting SFDA.
LogME is the baseline of our work.
Here, we describe the details for adapting these methods for object detection task.

{\bf KNAS} is originally used for Neural Architecture Search (NAS) under a gradient kernel hypothesis.
This hypothesis indicates that assuming $\boldsymbol{\mathcal{G}}$ is a set of all the gradients, there exists a gradient $\boldsymbol{g}$ which infers the downstream training performance.
We adopt it as a gradient based approach to compare with our gradient free approach.
Under this hypothesis, taking MSE loss for bounding box regression as an example, KNAS aims to minimize 
\begin{equation}
\small
\mathcal{L}(\boldsymbol{w})=\frac{1}{2}\left\|\hat{\boldsymbol{B}}-\boldsymbol{B}\right\|_2^2,
\label{equ:appendix_mse}
\end{equation}
where $\boldsymbol{w}$ is the trainable weights, $\hat{\boldsymbol{B}}\!=\![\hat{\boldsymbol{b}}_1, \dots, \hat{\boldsymbol{b}}_M]^T$ is the bounding box prediction matrix, $\boldsymbol{B}\!=\![\boldsymbol{b}_1, \dots, \boldsymbol{b}_M]^T$ is the ground truth bounding box matrix, and $M$ is the number of objects.
Then gradient descent is applied to optimize the model weights:
\begin{equation}
\small
\boldsymbol{\Theta}(t+1)=\boldsymbol{\Theta}(t)-\eta \frac{\partial \mathcal{L}(\boldsymbol{\Theta}(t))}{\partial \boldsymbol{\Theta}(t)},
\end{equation}
where $t$ represents the $t$-th iteration and $\eta$ is the learning rate.
The gradient for an object sample $i$ is 
\begin{equation}
\frac{\partial \mathcal{L}(\boldsymbol{\Theta}(t), i)}{\partial(\boldsymbol{\Theta}(t))}=\left(\hat{\boldsymbol{b}}_i-\boldsymbol{b}_i\right) \frac{\partial \hat{\boldsymbol{b}}_i}{\partial \boldsymbol{\Theta}(t)}.
\end{equation}
Then, a Gram matrix $\boldsymbol{H}$ is defined where the entry $(i, j)$ is 
\begin{equation}
\small
\boldsymbol{H}_{i, j}(t)=\left(\frac{\partial \hat{\boldsymbol{b}}_j(t)}{\partial \boldsymbol{\Theta}(t)}\right)\left(\frac{\partial \hat{\boldsymbol{b}}_i(t)}{\partial \boldsymbol{\Theta}(t)}\right)^T.
\end{equation}

$\boldsymbol{H}_{i, j}(t)$ is the dot-product between two gradient vectors $\boldsymbol{g}_i\!=\!\frac{\partial \hat{\boldsymbol{b}}_i(t)}{\partial \boldsymbol{\Theta}(t)}$ and $\boldsymbol{g}_j\!=\!\frac{\partial \hat{\boldsymbol{b}}_j(t)}{\partial \boldsymbol{\Theta}(t)}$.
To this end, the gradient kernel $\boldsymbol{g}$ can be computed as the mean of all elements in the Gram matrix $\boldsymbol{H}$:
\begin{equation}
\small
\boldsymbol{g}=\frac{1}{M^2} \sum_{i=1}^M\sum_{j=1}^M\left(\frac{\partial \hat{\boldsymbol{b}}_j(t)}{\partial \boldsymbol{\Theta}(t)}\right)\left(\frac{\partial \hat{\boldsymbol{b}}_i(t)}{\partial \boldsymbol{\Theta}(t)}\right)^T.
\label{equ:appendix_gradient_kernel}
\end{equation}
As the length of the whole gradient vector is too long, Eq. \eqref{equ:appendix_gradient_kernel} is approximated by 
\begin{equation}
\small
\boldsymbol{g}=\frac{1}{QM^2} \sum_{q=1}^Q\sum_{i=1}^M\sum_{j=1}^M\left(\frac{\partial \hat{\boldsymbol{b}}_j(t)}{\partial \hat{\boldsymbol{\Theta}}^q(t)}\right)\left(\frac{\partial \hat{\boldsymbol{b}}_i(t)}{\partial \hat{\boldsymbol{\Theta}}^q(t)}\right)^T.
\end{equation}
where $Q$ is the number of layers in the detection head, and $\hat{\boldsymbol{\Theta}}^q$ is the sampled parameters from $q$-th layer and the length of $\hat{\boldsymbol{\Theta}}^q$ is set as $1000$ in our implementation.
The obtained gradient kernel $\boldsymbol{g}$ is regarded as the transferability score from KNAS.

\begin{table*}[t]
\caption{Ranking results of of six methods for 1\% 33-choose-22 possible source model sets (over 1.9M) on $6$ downstream target datasets. Higher $\rho_w$ and Recall@1 indicate better ranking and transferability metric. As SFDA is specifically designed for classification task, it is not applicable for the single-class task of CrowdHuman. 
  The results of all three variants of our approach, U-LogME, IoU-LogME, and Det-LogME are reported. 
  The best methods are in {\color{red} red} and good ones are in {\color{blue} blue}.}
\vspace{-3mm}
\label{tab:appendix_pearson}%
  \centering
\scalebox{0.64}{

\begin{tabular}{l|cccccc|cccccc}
\toprule
Measure & \multicolumn{6}{c}{Weighted Pearson's Coefficient ($\rho_w$)} & \multicolumn{6}{c}{Recall@1} \\
Method & KNAS  & SFDA  & LogME & U-LogME & IoU-LogME & Det-LogME & KNAS  & SFDA  & LogME & U-LogME & IoU-LogME & Det-LogME \\
\midrule
Pascal VOC & 0.01$\pm$0.15 & \textcolor[rgb]{ 0,  0,  1}{\boldmath{}\textbf{0.71$\pm$0.14}\unboldmath{}} & -0.04$\pm$0.16 & -0.07$\pm$0.23 & \textcolor[rgb]{ 1,  0,  0}{\boldmath{}\textbf{0.73$\pm$0.13}\unboldmath{}} & 0.68$\pm$0.12 & 0.26$\pm$0.44 & 0.33$\pm$0.47 & \textcolor[rgb]{ 1,  0,  0}{\boldmath{}\textbf{0.53$\pm$0.50}\unboldmath{}} & 0.20$\pm$0.40 & 0.34$\pm$0.47 & \textcolor[rgb]{ 0,  0,  1}{\boldmath{}\textbf{0.41$\pm$0.49}\unboldmath{}} \\
CityScapes & 0.15$\pm$0.18 & 0.46$\pm$0.11 & 0.38$\pm$0.09 & 0.19$\pm$0.13 & \textcolor[rgb]{ 0,  0,  1}{\boldmath{}\textbf{0.53$\pm$0.10}\unboldmath{}} & \textcolor[rgb]{ 1,  0,  0}{\boldmath{}\textbf{0.55$\pm$0.09}\unboldmath{}} & \textcolor[rgb]{ 1,  0,  0}{\boldmath{}\textbf{0.53$\pm$0.50}\unboldmath{}} & 0.00$\pm$0.00 & \textcolor[rgb]{ 1,  0,  0}{\boldmath{}\textbf{0.53$\pm$0.50}\unboldmath{}} & 0.12$\pm$0.33 & \textcolor[rgb]{ 1,  0,  0}{\boldmath{}\textbf{0.53$\pm$0.50}\unboldmath{}} & \textcolor[rgb]{ 1,  0,  0}{\boldmath{}\textbf{0.53$\pm$0.50}\unboldmath{}} \\
SODA  & -0.11$\pm$0.21 & 0.60$\pm$0.13 & 0.28$\pm$0.13 & 0.12$\pm$0.13 & \textcolor[rgb]{ 0,  0,  1}{\boldmath{}\textbf{0.65$\pm$0.12}\unboldmath{}} & \textcolor[rgb]{ 1,  0,  0}{\boldmath{}\textbf{0.66$\pm$0.11}\unboldmath{}} & 0.00$\pm$0.00 & 0.00$\pm$0.00 & \textcolor[rgb]{ 1,  0,  0}{\boldmath{}\textbf{0.53$\pm$0.50}\unboldmath{}} & 0.12$\pm$0.33 & \textcolor[rgb]{ 1,  0,  0}{\boldmath{}\textbf{0.53$\pm$0.50}\unboldmath{}} & \textcolor[rgb]{ 1,  0,  0}{\boldmath{}\textbf{0.53$\pm$0.50}\unboldmath{}} \\
CrowdHuman & -0.21$\pm$0.13 & N/A   & 0.08$\pm$0.19 & 0.11$\pm$0.17 & \textcolor[rgb]{ 1,  0,  0}{\boldmath{}\textbf{0.31$\pm$0.08}\unboldmath{}} & \textcolor[rgb]{ 0,  0,  1}{\boldmath{}\textbf{0.30$\pm$0.08}\unboldmath{}} & 0.00$\pm$0.00 & N/A   & \textcolor[rgb]{ 1,  0,  0}{\boldmath{}\textbf{0.65$\pm$0.48}\unboldmath{}} & 0.58$\pm$0.49 & \textcolor[rgb]{ 1,  0,  0}{\boldmath{}\textbf{0.65$\pm$0.48}\unboldmath{}} & \textcolor[rgb]{ 1,  0,  0}{\boldmath{}\textbf{0.65$\pm$0.48}\unboldmath{}} \\
VisDrone & 0.15$\pm$0.21 & 0.29$\pm$0.15 & 0.35$\pm$0.10 & 0.12$\pm$0.10 & \textcolor[rgb]{ 1,  0,  0}{\boldmath{}\textbf{0.44$\pm$0.12}\unboldmath{}} & \textcolor[rgb]{ 1,  0,  0}{\boldmath{}\textbf{0.44$\pm$0.11}\unboldmath{}} & 0.12$\pm$0.32 & 0.34$\pm$0.47 & 0.17$\pm$0.38 & 0.01$\pm$0.11 & \textcolor[rgb]{ 1,  0,  0}{\boldmath{}\textbf{0.25$\pm$0.43}\unboldmath{}} & \textcolor[rgb]{ 1,  0,  0}{\boldmath{}\textbf{0.25$\pm$0.43}\unboldmath{}} \\
DeepLesion & 0.08$\pm$0.18 & -0.37$\pm$0.29 & 0.34$\pm$0.20 & \textcolor[rgb]{ 1,  0,  0}{\boldmath{}\textbf{0.54$\pm$0.19}\unboldmath{}} & -0.17$\pm$0.34 & \textcolor[rgb]{ 0,  0,  1}{\boldmath{}\textbf{0.50$\pm$0.16}\unboldmath{}} & 0.01$\pm$0.09 & 0.00$\pm$0.00 & 0.26$\pm$0.44 & \textcolor[rgb]{ 1,  0,  0}{\boldmath{}\textbf{0.57$\pm$0.50}\unboldmath{}} & 0.00$\pm$0.03 & \textcolor[rgb]{ 0,  0,  1}{\boldmath{}\textbf{0.42$\pm$0.49}\unboldmath{}} \\
\midrule
Average & 0.01$\pm$0.18 & 0.34$\pm$0.16 & 0.23$\pm$0.15 & 0.20$\pm$0.16 & \textcolor[rgb]{ 0,  0,  1}{\boldmath{}\textbf{0.42$\pm$0.15}\unboldmath{}} & \textcolor[rgb]{ 1,  0,  0}{\boldmath{}\textbf{0.52$\pm$0.11}\unboldmath{}} & 0.15$\pm$0.36 & 0.11$\pm$0.31 & \textcolor[rgb]{ 0,  0,  1}{\boldmath{}\textbf{0.44$\pm$0.50}\unboldmath{}} & 0.27$\pm$0.44 & 0.38$\pm$0.49 & \textcolor[rgb]{ 1,  0,  0}{\boldmath{}\textbf{0.46$\pm$0.50}\unboldmath{}} \\
\bottomrule
\end{tabular}%

}
\end{table*}%

\begin{table*}[t]
  \caption{The transferability scores obtained from $6$ metrics and fine-tuning mAP on Pascal VOC and CityScapes datasets. The last row is the corresponding ranking correlation $\tau_w$ for every metric.}
\vspace{-3mm}
  \label{tab:pascal_voc_cityscapes}%
  \centering
  \scalebox{0.57}{

\begin{tabular}{llccccccc|ccccccc}
\toprule
\multirow{2}[4]{*}{Model} & \multirow{2}[4]{*}{Backbone} & \multicolumn{7}{c|}{Pascal VOC}                       & \multicolumn{7}{c}{CityScapes} \\
\cmidrule{3-16}      &       & KNAS  & SFDA  & LogME & U-LogME & IoU-LogME & Det-LogME & mAP   & KNAS  & SFDA  & LogME & U-LogME & IoU-LogME & Det-LogME & mAP \\
\midrule
\multirow{4}[2]{*}{Faster RCNN} & R50   & 2.326E-01 & 0.791  & -6.193  & -3.223  & 0.482  & 1.199  & 84.5  & -2.093E+00 & 0.879  & -6.257  & -1.518  & 0.624  & 9.229  & 41.9  \\
      & R101  & 1.095E-01 & 0.809  & -6.177  & -3.160  & 0.492  & 1.258  & 84.5  & -1.791E+00 & 0.887  & -6.258  & -1.478  & 0.624  & 9.289  & 42.3  \\
      & X101-32x4d & -4.396E-02 & 0.822  & -6.146  & -2.969  & 0.505  & 1.380  & 85.2  & -2.386E+00 & 0.892  & -6.269  & -1.397  & 0.622  & 9.242  & 43.5  \\
      & X101-64x4d & 9.018E-01 & 0.825  & -6.129  & -2.944  & 0.509  & 1.405  & 85.6  & -1.664E+00 & 0.894  & -6.270  & -1.381  & 0.624  & 9.353  & 42.8  \\
\cmidrule{2-16}\multirow{4}[2]{*}{Cascade RCNN} & R50   & -6.438E-01 & 0.795  & -6.232  & -3.203  & 0.481  & 1.206  & 84.1  & -6.218E+00 & 0.874  & -6.286  & -1.514  & 0.618  & 9.013  & 44.1  \\
      & R101  & -2.226E-01 & 0.811  & -6.222  & -3.176  & 0.490  & 1.247  & 84.9  & -6.553E+00 & 0.883  & -6.289  & -1.489  & 0.621  & 9.127  & 43.7  \\
      & X101-32x4d & -6.405E-01 & 0.826  & -6.194  & -3.024  & 0.503  & 1.351  & 85.6  & -5.763E+00 & 0.891  & -6.297  & -1.415  & 0.620  & 9.160  & 44.1  \\
      & X101-64x4d & 1.270E+00 & 0.831  & -6.190  & -3.006  & 0.505  & 1.367  & 85.8  & -5.182E+00 & 0.891  & -6.290  & -1.402  & 0.620  & 9.166  & 45.4  \\
\cmidrule{2-16}Dynamic RCNN & R50   & -8.148E-03 & 0.791  & -6.206  & -2.875  & 0.483  & 1.343  & 84.0  & 1.878E-01 & 0.869  & -6.303  & -1.352  & 0.617  & 9.110  & 42.5  \\
\cmidrule{2-16}\multirow{5}[2]{*}{RegNet} & 400MF & 5.056E-01 & 0.750  & -6.162  & -3.387  & 0.465  & 1.076  & 83.3  & 2.401E-01 & 0.845  & -6.264  & -1.647  & 0.606  & 8.400  & 39.9  \\
      & 800MF & 6.691E-02 & 0.758  & -6.156  & -3.295  & 0.468  & 1.122  & 83.9  & -2.308E+00 & 0.855  & -6.279  & -1.606  & 0.606  & 8.441  & 40.3  \\
      & 1.6GF & 1.523E-01 & 0.770  & -6.162  & -3.232  & 0.472  & 1.161  & 84.6  & -1.504E+00 & 0.869  & -6.279  & -1.553  & 0.613  & 8.767  & 41.8  \\
      & 3.2GF & 6.241E-02 & 0.786  & -6.170  & -3.186  & 0.482  & 1.215  & 85.5  & -3.148E-01 & 0.877  & -6.269  & -1.527  & 0.618  & 8.984  & 42.7  \\
      & 4GF   & 3.995E-01 & 0.790  & -6.166  & -3.133  & 0.484  & 1.242  & 85.0  & -1.451E+00 & 0.878  & -6.278  & -1.506  & 0.617  & 8.956  & 43.1  \\
\cmidrule{2-16}\multirow{3}[2]{*}{DCN} & R50   & 3.852E-02 & 0.825  & -6.122  & -2.748  & 0.511  & 1.490  & 86.1  & -6.647E-01 & 0.889  & -6.246  & -1.267  & 0.625  & 9.497  & 42.6  \\
      & R101  & -9.254E-02 & 0.836  & -6.155  & -2.812  & 0.516  & 1.481  & 86.5  & -2.072E+00 & 0.894  & -6.253  & -1.298  & 0.626  & 9.503  & 43.1  \\
      & X101-32x4d & 7.048E-02 & 0.846  & -6.100  & -2.653  & 0.525  & 1.577  & 86.9  & -7.308E-01 & 0.899  & -6.253  & -1.227  & 0.626  & 9.571  & 43.5  \\
\midrule
\multirow{2}[2]{*}{FCOS} & R50   & 1.023E+01 & 0.289  & -6.093  & -1.856  & 0.264  & 0.988  & 77.3  & 6.343E+00 & 0.492  & -6.434  & -0.992  & 0.491  & 4.318  & 40.4  \\
      & R101  & 5.233E+00 & 0.280  & -6.032  & -2.101  & 0.262  & 0.884  & 79.4  & 5.277E+00 & 0.515  & -6.426  & -1.124  & 0.491  & 4.219  & 41.2  \\
\cmidrule{2-16}\multirow{5}[2]{*}{RetinaNet} & R18   & -3.404E-01 & 0.733  & -6.289  & -2.928  & 0.442  & 1.177  & 80.9  & 1.157E-02 & 0.844  & -6.411  & -1.438  & 0.597  & 8.206  & 36.7  \\
      & R50   & -1.357E-01 & 0.759  & -6.277  & -2.975  & 0.457  & 1.213  & 84.1  & 5.439E-03 & 0.867  & -6.370  & -1.388  & 0.606  & 8.609  & 40.0  \\
      & R101  & -1.807E-01 & 0.774  & -6.259  & -2.972  & 0.467  & 1.246  & 84.4  & 4.709E-02 & 0.879  & -6.357  & -1.374  & 0.612  & 8.854  & 40.6  \\
      & X101-32x4d & -1.030E-01 & 0.792  & -6.260  & -2.763  & 0.475  & 1.360  & 84.6  & 2.935E-02 & 0.881  & -6.377  & -1.308  & 0.608  & 8.762  & 41.2  \\
      & X101-64x4d & -3.170E-01 & 0.792  & -6.229  & -2.722  & 0.475  & 1.376  & 85.3  & 2.304E-02 & 0.886  & -6.366  & -1.276  & 0.610  & 8.858  & 42.0  \\
\cmidrule{2-16}\multirow{2}[2]{*}{Sparse RCNN} & R50   & 9.846E+03 & 0.777  & -6.267  & -3.243  & 0.456  & 1.102  & 84.7  & -1.640E+04 & 0.878  & -6.414  & -1.595  & 0.602  & 8.304  & 38.9  \\
      & R101  & -2.104E+04 & 0.795  & -6.238  & -3.263  & 0.466  & 1.127  & 85.0  & 1.198E+04 & 0.884  & -6.396  & -1.601  & 0.602  & 8.304  & 39.3  \\
\midrule
Deformable DETR & R50   & 8.873E+03 & 0.794  & -5.221  & -2.295  & 0.462  & 1.501  & 87.0  & 1.363E+05 & 0.881  & -5.376  & -1.065  & 0.673  & 11.602  & 45.5  \\
\midrule
Faster RCNN OI & R50   & 2.038E+00 & 0.724  & -6.016  & -4.100  & 0.443  & 0.716  & 82.2  & 3.288E+00 & 0.837  & -6.260  & -1.951  & 0.602  & 7.982  & 39.3  \\
RetinaNet OI & R50   & -2.045E-01 & 0.697  & -6.195  & -3.335  & 0.430  & 0.974  & 82.0  & 1.701E-01 & 0.845  & -6.343  & -1.624  & 0.600  & 8.177  & 39.5  \\
\midrule
SoCo  & R50   & -3.222E+00 & 0.703  & -6.094  & -3.062  & 0.433  & 1.093  & 56.5  & 4.629E+01 & 0.836  & -6.237  & -1.473  & 0.606  & 8.536  & 41.7  \\
InsLoc & R50   & -3.153E-03 & 0.566  & -6.239  & -1.592  & 0.424  & 1.649  & 86.7  & 1.041E-02 & 0.756  & -6.322  & -0.738  & 0.582  & 8.191  & 40.3  \\
\midrule
UP-DETR & R50   & 8.225E+02 & 0.175  & -6.267  & -3.086  & 0.238  & 0.404  & 59.3  & -2.994E+02 & 0.399  & -6.485  & -1.404  & 0.403  & 0.455  & 30.9  \\
DETReg & R50   & -6.832E+02 & 0.189  & -5.999  & -3.872  & 0.248  & 0.129  & 63.5  & -9.335E+02 & 0.427  & -5.892  & -1.958  & 0.440  & 1.462  & 38.7  \\
\midrule
\multicolumn{2}{l}{$\tau_w$} & 0.15  & 0.64  & 0.22  & 0.43  & 0.54  & \textbf{0.79 } & N/A   & -0.02  & 0.51  & 0.32  & 0.18  & 0.68  & \textbf{0.71 } & N/A \\
\bottomrule
\end{tabular}%

}

\end{table*}%

\begin{table*}[t]
  \caption{The transferability scores obtained from $6$ metrics and fine-tuning mAP on SODA and CrowdHuman datasets. The last row is the corresponding ranking correlation $\tau_w$ for every metric.}
\vspace{-3mm}
  \label{tab:soda_crowdhuman}
  \centering
  \scalebox{0.57}{

\begin{tabular}{llccccccc|ccccccc}
\toprule
\multirow{2}[4]{*}{Model} & \multirow{2}[4]{*}{Backbone} & \multicolumn{7}{c|}{SODA}                             & \multicolumn{7}{c}{CrowdHuman} \\
\cmidrule{3-16}      &       & KNAS  & SFDA  & LogME & U-LogME & IoU-LogME & Det-LogME & mAP   & KNAS  & SFDA  & LogME & U-LogME & IoU-LogME & Det-LogME & mAP \\
\midrule
\multirow{4}[2]{*}{Faster RCNN} & R50   & -1.314E+00 & 0.831  & -5.698  & -2.148  & 0.542  & 16.546  & 34.7  & -1.216E+01 & N/A   & -6.660  & -0.116  & 0.575  & 1.357  & 41.4  \\
      & R101  & -2.636E+00 & 0.846  & -5.679  & -2.071  & 0.548  & 17.121  & 35.0  & -1.648E+01 & N/A   & -6.655  & -0.111  & 0.577  & 1.482  & 41.3  \\
      & X101-32x4d & -2.516E+00 & 0.852  & -5.664  & -1.924  & 0.554  & 17.643  & 35.7  & -7.494E+00 & N/A   & -6.620  & -0.074  & 0.586  & 2.081  & 41.2  \\
      & X101-64x4d & -1.226E+00 & 0.856  & -5.660  & -1.914  & 0.557  & 17.900  & 36.4  & -9.416E+00 & N/A   & -6.596  & -0.045  & 0.592  & 2.460  & 41.5  \\
\cmidrule{2-16}\multirow{4}[2]{*}{Cascade RCNN} & R50   & -9.109E+00 & 0.826  & -5.746  & -2.129  & 0.537  & 16.183  & 35.3  & -1.958E+01 & N/A   & -6.681  & -0.138  & 0.568  & 0.851  & 43.0  \\
      & R101  & -1.177E+01 & 0.836  & -5.733  & -2.091  & 0.543  & 16.685  & 35.9  & -2.723E+01 & N/A   & -6.683  & -0.143  & 0.567  & 0.781  & 42.8  \\
      & X101-32x4d & -1.509E+01 & 0.846  & -5.709  & -1.966  & 0.549  & 17.245  & 36.8  & -1.755E+01 & N/A   & -6.662  & -0.126  & 0.573  & 1.170  & 43.2  \\
      & X101-64x4d & -1.277E+01 & 0.851  & -5.733  & -1.946  & 0.552  & 17.453  & 37.4  & -1.083E+01 & N/A   & -6.660  & -0.121  & 0.573  & 1.225  & 43.7  \\
\cmidrule{2-16}Dynamic RCNN & R50   & -1.597E+00 & 0.820  & -5.758  & -1.871  & 0.535  & 16.169  & 35.2  & -5.448E+00 & N/A   & -6.674  & -0.130  & 0.568  & 0.900  & 41.8  \\
\cmidrule{2-16}\multirow{5}[2]{*}{RegNet} & 400MF & -1.900E+00 & 0.785  & -5.670  & -2.300  & 0.520  & 14.767  & 32.5  & -1.393E+01 & N/A   & -6.628  & -0.099  & 0.579  & 1.619  & 38.0  \\
      & 800MF & -1.512E+00 & 0.803  & -5.694  & -2.239  & 0.525  & 15.206  & 34.2  & -8.908E+00 & N/A   & -6.636  & -0.099  & 0.578  & 1.546  & 39.8  \\
      & 1.6GF & -2.576E+00 & 0.815  & -5.668  & -2.169  & 0.537  & 16.190  & 35.7  & -1.558E+01 & N/A   & -6.653  & -0.118  & 0.573  & 1.215  & 41.8  \\
      & 3.2GF & -2.007E+00 & 0.827  & -5.682  & -2.140  & 0.538  & 16.288  & 37.0  & -1.560E+01 & N/A   & -6.647  & -0.106  & 0.577  & 1.438  & 41.7  \\
      & 4GF   & -1.735E+00 & 0.826  & -5.700  & -2.097  & 0.539  & 16.380  & 37.0  & -1.600E+01 & N/A   & -6.650  & -0.111  & 0.576  & 1.388  & 41.9  \\
\cmidrule{2-16}\multirow{3}[1]{*}{DCN} & R50   & -1.077E+00 & 0.844  & -5.677  & -1.736  & 0.553  & 17.632  & 35.3  & -1.562E+01 & N/A   & -6.569  & -0.013  & 0.602  & 3.121  & 43.1  \\
      & R101  & -8.408E-01 & 0.846  & -5.669  & -1.786  & 0.556  & 17.874  & 35.3  & -1.073E+01 & N/A   & -6.584  & -0.033  & 0.596  & 2.718  & 43.4  \\
      & X101-32x4d & -1.797E+00 & 0.859  & -5.676  & -1.655  & 0.559  & 18.189  & 36.0  & -7.181E+00 & N/A   & -6.494  & 0.018  & 0.614  & 3.876  & 44.3  \\
      \midrule
\multirow{2}[1]{*}{FCOS} & R50   & -1.287E-02 & 0.510  & -5.729  & -1.328  & 0.415  & 6.902  & 33.3  & -1.263E+00 & N/A   & -6.578  & -0.040  & 0.570  & 1.046  & 35.6  \\
      & R101  & 6.980E-01 & 0.541  & -5.688  & -1.535  & 0.413  & 6.610  & 34.5  & -1.601E+00 & N/A   & -6.537  & -0.006  & 0.578  & 1.593  & 36.8  \\
\cmidrule{2-16}\multirow{5}[2]{*}{RetinaNet} & R18   & -2.071E-02 & 0.778  & -5.846  & -1.988  & 0.510  & 14.099  & 29.6  & 3.133E-02 & N/A   & -6.696  & -0.161  & 0.554  & 0.008  & 35.8  \\
      & R50   & -7.809E-01 & 0.818  & -5.817  & -1.885  & 0.527  & 15.528  & 33.9  & 3.080E-02 & N/A   & -6.702  & -0.168  & 0.555  & 0.035  & 38.3  \\
      & R101  & -1.857E-02 & 0.828  & -5.835  & -1.874  & 0.532  & 15.877  & 34.0  & -1.330E-03 & N/A   & -6.699  & -0.164  & 0.556  & 0.121  & 38.6  \\
      & X101-32x4d & -1.033E-01 & 0.833  & -5.864  & -1.794  & 0.530  & 15.766  & 34.2  & -4.154E-03 & N/A   & -6.691  & -0.157  & 0.557  & 0.171  & 38.9  \\
      & X101-64x4d & -2.995E-01 & 0.839  & -5.851  & -1.717  & 0.537  & 16.430  & 35.6  & 8.897E-03 & N/A   & -6.676  & -0.147  & 0.562  & 0.477  & 39.9  \\
\cmidrule{2-16}\multirow{2}[2]{*}{Sparse RCNN} & R50   & -5.855E+04 & 0.824  & -5.892  & -2.256  & 0.518  & 14.636  & 35.9  & 4.174E+04 & N/A   & -6.683  & -0.154  & 0.554  & 0.009  & 38.6  \\
      & R101  & -1.934E+05 & 0.833  & -5.869  & -2.255  & 0.523  & 14.991  & 36.3  & -1.077E+03 & N/A   & -6.676  & -0.145  & 0.557  & 0.190  & 39.2  \\
\midrule
Deformable DETR & R50   & -9.691E+04 & 0.820  & -4.557  & -1.421  & 0.589  & 20.697  & 38.8  & -3.523E+04 & N/A   & -5.447  & 0.904  & 0.790  & 15.536  & 45.3  \\
\midrule
Faster RCNN OI & R50   & -2.170E+01 & 0.767  & -5.543  & -2.756  & 0.513  & 13.942  & 32.8  & -6.768E+01 & N/A   & -6.533  & -0.006  & 0.601  & 3.037  & 40.0  \\
RetinaNet OI & R50   & -2.793E-01 & 0.780  & -5.782  & -2.278  & 0.507  & 13.741  & 33.4  & 7.107E-03 & N/A   & -6.650  & -0.104  & 0.571  & 1.077  & 38.5  \\
\midrule
SoCo  & R50   & -5.681E-01 & 0.759  & -5.584  & -2.074  & 0.517  & 14.607  & 33.2  & -1.681E+01 & N/A   & -6.553  & 0.019  & 0.601  & 3.065  & 40.6  \\
InsLoc & R50   & 1.857E-02 & 0.691  & -5.750  & -0.842  & 0.490  & 13.092  & 31.4  & 7.323E-01 & N/A   & -6.652  & -0.117  & 0.565  & 0.690  & 40.7  \\
\midrule
UP-DETR & R50   & -4.658E+02 & 0.457  & -6.040  & -1.946  & 0.338  & 0.423  & 20.1  & 3.661E+02 & N/A   & -6.613  & -0.145  & 0.555  & 0.062  & 35.4  \\
DETReg & R50   & -8.563E+02 & 0.467  & -5.584  & -2.477  & 0.371  & 2.783  & 24.3  & -6.686E+02 & N/A   & -6.202  & 0.221  & 0.638  & 5.498  & 41.0  \\
\midrule
\multicolumn{2}{l}{$\tau_w$} & -0.44  & 0.43  & 0.22  & 0.03  & \textbf{0.66 } & 0.65  & N/A   & -0.47  & N/A   & 0.37  & 0.39  & \textbf{0.51 } & \textbf{0.51 } & N/A \\
\bottomrule
\end{tabular}%

}

\vspace{-3mm}

\end{table*}%

{\bf SFDA} is specially designed to assess the transferability for classification tasks, which is not applicable for single-class detection datasets used in this work including SKU-110K \cite{goldman2019precise}, WIDER FACE \cite{yang2016wider}, and CrowdHuman \cite{shao2018crowdhuman}.
It aims to leverage the neglected fine-tuning dynamics for transferability evaluation, which degrades the efficiency.
Given object-level feature matrix $\boldsymbol{F}\!=\![\boldsymbol{f}_1, \dots, \boldsymbol{f}_M]^T$, with corresponding class label matrix $\boldsymbol{C}$, we consider object detection as an object-level multi-class classification task for adapting SFDA.

To utilize the fine-tuning dynamics, SFDA transforms the object feature matrix $\boldsymbol{F}$ to a space with good class separation under Regularized Fisher Discriminant Analysis (Reg-FDA).
A transformation is defined to project $\boldsymbol{F}\!\!\in\!\!\mathbb{R}^{M \times D}$ to $\tilde{\boldsymbol{F}}\!\!\in\!\!\mathbb{R}^{M \times D'}$ by a projection matrix $\boldsymbol{U}\!\!\in\!\!\mathbb{R}^{D \times D'}$ with $\tilde{\boldsymbol{F}}:=\boldsymbol{U}^T\boldsymbol{F}$.
The project matrix is 
\begin{equation}
\small
\boldsymbol{U}=\arg\max_{\boldsymbol{U}} \frac{d_b(\boldsymbol{U})}{d_w(\boldsymbol{U})} \stackrel{\text { def }}{=} \frac{\left|\boldsymbol{U}^{\top} \boldsymbol{S}_b\boldsymbol{U}\right|}{\left|\boldsymbol{U}^{\top}\left[(1-\lambda) \boldsymbol{S}_w+\lambda \boldsymbol{I}\right] \boldsymbol{U}\right|},
\end{equation}
where $d_b(\boldsymbol{U})$ and $d_w(\boldsymbol{U})$ represent between scatter of classes and within scatter of each class, $\lambda\!\!\in\!\![0, 1]$ is a regularization coefficient for a trade-off between the inter-class separation and intra-class compactness, and $\boldsymbol{I}$ is an identity matrix.
The between and within scatter matrix $\boldsymbol{S}_b$ and $\boldsymbol{S}_w$ are difined as
\begin{equation}
\small
\begin{aligned}
\centering
\boldsymbol{S}_b&=\sum_{c=1}^K M_c\left(\nu_c-\nu\right)\left(\nu_c-\nu\right)^{\top}\\
\boldsymbol{S}_w&=\sum_{c=1}^K \sum_{i=1}^{M_c}\left(\boldsymbol{f}_i^{(c)}-\nu_c\right)\left(\boldsymbol{f}_i^{(c)}-\nu_c\right)^{\top},
\end{aligned}
\end{equation}
where $\nu=\sum_{i=1}^M \boldsymbol{f}_i$ and $\nu_c=\sum_{i=1}^M \boldsymbol{f}_i^{(c)}$ are the mean of all and $c$-th class object features.

With the intuition that a model with Infomin requires stronger supervision for minimizing within scatter of every class which results in better classes separation.
$\lambda$ is instantiated by $\lambda=\exp ^{-a \sigma\left(\boldsymbol{S}_w\right)}$, where $a$ is a positive constant and $\sigma(\boldsymbol{S}_w)$ is the largest eigenvalue of $\boldsymbol{S}_w$.
For every class, SFDA assumes $\tilde{\boldsymbol{f}}_i^{(c)} \sim \mathcal{N}\left(\boldsymbol{U}^{\top} \nu_c, \Sigma_c\right)$, where $\Sigma_c$ is the covariance matrix of $\{\tilde{\boldsymbol{f}}_i^{(c)}\}_{i=1}^{M_c}$.
With projection matrix $\boldsymbol{U}$, the score function for class $c$ is 
\begin{equation}
\small
\delta_c\left(\boldsymbol{f}_i\right)=\boldsymbol{f}_i^{\top} \boldsymbol{U}\boldsymbol{U}^{\top} \nu_c-\frac{1}{2} \nu_c^{\top} \boldsymbol{U} \boldsymbol{U}^{\top} \nu_c+\log \frac{M_c}{M}.
\end{equation}
Then, the final class prediction probability is obtained by normalizing $\{\delta_c\left(\boldsymbol{f}_i\right)\}_K$ with softmax function:
\begin{equation}
\small
p\left(c_i\!\mid\!\boldsymbol{f}_i\right)=\frac{\exp ^{\delta_{c_i}\left(\boldsymbol{f}_i\right)}}{\sum_{c=1}^K \exp ^{\delta_c\left(\boldsymbol{f}_i\right)}}
\end{equation}
To this end, the transferability score is expressed as the mean of $p\left(c_i\!\mid\!\boldsymbol{f}_i\right)$ over all object samples by
\begin{equation}
\small
p\left(\boldsymbol{C}\!\mid\!\boldsymbol{F}\right)=\frac{1}{M}\sum_{i=1}^M\frac{\exp ^{\delta_{c_i}\left(\boldsymbol{f}_i\right)}}{\sum_{c=1}^K \exp ^{\delta_c\left(\boldsymbol{f}_i\right)}}.
\end{equation}

{\bf LogME} is following Eq. \eqref{equ:appendix_logme} described in Sec. \ref{sec:appendix_logme}.

\begin{table*}[htbp]
  \caption{The transferability scores obtained from $6$ metrics and fine-tuning mAP on VisDrone and DeepLesion datasets. The last row is the corresponding ranking correlation $\tau_w$ for every metric.}
\vspace{-3mm}
  \label{tab:visdrone_deeplesion}%
  \centering
\scalebox{0.57}{

\begin{tabular}{llccccccc|ccccccc}
\toprule
\multirow{2}[3]{*}{Model} & \multirow{2}[3]{*}{Backbone} & \multicolumn{7}{c|}{VisDrone}                         & \multicolumn{7}{c}{DeepLesion} \\
\cmidrule{3-16}      &       & KNAS  & SFDA  & LogME & U-LogME & IoU-LogME & Det-LogME & mAP   & KNAS  & SFDA  & LogME & U-LogME & IoU-LogME & Det-LogME & mAP \\
\multirow{4}[1]{*}{Faster RCNN} & R50   & 3.634E-01 & 0.645  & -6.611  & -1.771  & 0.432  & 1.357  & 21.3  & 6.270E-01 & 0.665  & -4.858  & -3.203  & 0.394  & 1.114  & 2.9  \\
      & R101  & -3.928E-01 & 0.662  & -6.608  & -1.734  & 0.438  & 1.482  & 21.5  & 6.524E-01 & 0.675  & -4.801  & -3.118  & 0.396  & 1.129  & 2.8  \\
      & X101-32x4d & -1.176E+00 & 0.661  & -6.551  & -1.648  & 0.442  & 2.081  & 22.3  & 4.738E-01 & 0.691  & -4.774  & -2.838  & 0.392  & 1.137  & 3.0  \\
      & X101-64x4d & -5.828E-01 & 0.669  & -6.527  & -1.636  & 0.443  & 2.460  & 23.2  & 3.634E-01 & 0.682  & -4.778  & -2.967  & 0.386  & 1.107  & 3.7  \\
\cmidrule{2-16}\multirow{4}[2]{*}{Cascade RCNN} & R50   & -8.062E-01 & 0.637  & -6.652  & -1.775  & 0.427  & 0.851  & 20.7  & -1.159E+00 & 0.662  & -4.795  & -3.041  & 0.387  & 1.102  & 3.1  \\
      & R101  & -2.153E+00 & 0.645  & -6.649  & -1.764  & 0.428  & 0.781  & 21.4  & -1.697E+00 & 0.674  & -4.752  & -3.295  & 0.392  & 1.100  & 3.1  \\
      & X101-32x4d & -2.963E+00 & 0.659  & -6.607  & -1.687  & 0.436  & 1.170  & 21.9  & -7.776E-01 & 0.685  & -4.717  & -2.981  & 0.395  & 1.133  & 3.6  \\
      & X101-64x4d & -4.002E+00 & 0.662  & -6.600  & -1.670  & 0.438  & 1.225  & 22.5  & -1.154E+00 & 0.678  & -4.667  & -3.202  & 0.385  & 1.083  & 3.0  \\
\cmidrule{2-16}Dynamic RCNN & R50   & 1.827E-01 & 0.639  & -6.629  & -1.650  & 0.433  & 0.900  & 16.1  & 9.446E-01 & 0.652  & -4.712  & -1.752  & 0.396  & 1.237  & 3.0  \\
\cmidrule{2-16}\multirow{5}[2]{*}{RegNet} & 400MF & -9.387E-01 & 0.609  & -6.497  & -1.902  & 0.433  & 1.619  & 19.2  & 7.476E-01 & 0.660  & -4.826  & -2.895  & 0.392  & 1.131  & 2.8  \\
      & 800MF & -6.771E-01 & 0.631  & -6.552  & -1.860  & 0.437  & 1.546  & 21.1  & 1.511E-01 & 0.641  & -4.871  & -2.528  & 0.392  & 1.162  & 2.9  \\
      & 1.6GF & -2.191E-01 & 0.646  & -6.588  & -1.831  & 0.438  & 1.215  & 22.2  & 5.915E-01 & 0.666  & -4.770  & -2.723  & 0.403  & 1.183  & 3.1  \\
      & 3.2GF & -1.319E+00 & 0.657  & -6.584  & -1.801  & 0.442  & 1.438  & 23.3  & 4.484E-01 & 0.658  & -4.790  & -2.486  & 0.397  & 1.182  & 3.3  \\
      & 4GF   & -2.118E+00 & 0.654  & -6.572  & -1.785  & 0.442  & 1.388  & 23.2  & 1.133E+00 & 0.642  & -4.833  & -2.539  & 0.396  & 1.173  & 2.8  \\
\cmidrule{2-16}\multirow{3}[2]{*}{DCN} & R50   & -1.394E+00 & 0.654  & -6.513  & -1.582  & 0.447  & 3.121  & 21.7  & 3.988E-01 & 0.705  & -4.570  & -2.418  & 0.425  & 1.282  & 2.7  \\
      & R101  & -1.431E+00 & 0.666  & -6.529  & -1.623  & 0.447  & 2.718  & 21.9  & -7.920E-01 & 0.707  & -4.602  & -2.527  & 0.431  & 1.293  & 3.0  \\
      & X101-32x4d & -3.897E-01 & 0.677  & -6.397  & -1.537  & 0.458  & 3.876  & 23.3  & 3.103E-01 & 0.698  & -4.573  & -2.023  & 0.421  & 1.300  & 3.5  \\
\midrule
\multirow{2}[2]{*}{FCOS} & R50   & 5.389E+00 & 0.476  & -6.523  & -1.362  & 0.393  & 1.046  & \multicolumn{1}{c}{21.6 } & -5.430E+00 & 0.254  & -4.747  & 6.207  & 0.295  & 1.542  & 4.5  \\
      & R101  & 5.258E+00 & 0.493  & -6.448  & -1.474  & 0.396  & 1.593  & \multicolumn{1}{c}{22.4 } & -6.504E+00 & 0.202  & -4.352  & 5.016  & 0.283  & 1.405  & 4.8  \\
\cmidrule{2-16}\multirow{5}[2]{*}{RetinaNet} & R18   & -4.476E-02 & 0.603  & -6.687  & -1.712  & 0.419  & 0.008  & 14.7  & -4.071E-02 & 0.525  & -4.836  & -1.504  & 0.410  & 1.306  & 2.8  \\
      & R50   & -1.586E-02 & 0.645  & -6.695  & -1.644  & 0.427  & 0.035  & 17.9  & -1.368E-02 & 0.513  & -4.871  & -1.081  & 0.389  & 1.268  & 3.4  \\
      & R101  & -8.479E-02 & 0.650  & -6.678  & -1.637  & 0.430  & 0.121  & 18.2  & 1.456E-01 & 0.569  & -4.822  & -1.087  & 0.416  & 1.360  & 3.7  \\
      & X101-32x4d & -2.029E-01 & 0.647  & -6.681  & -1.600  & 0.427  & 0.171  & 18.5  & 6.097E-01 & 0.515  & -4.824  & -0.987  & 0.412  & 1.355  & 4.5  \\
      & X101-64x4d & -9.568E-02 & 0.662  & -6.645  & -1.538  & 0.434  & 0.477  & 19.1  & 1.775E-01 & 0.541  & -4.857  & -0.628  & 0.412  & 1.383  & 4.2  \\
\cmidrule{2-16}\multirow{2}[1]{*}{Sparse RCNN} & R50   & -2.382E+03 & 0.643  & -6.695  & -1.887  & 0.425  & 0.009  & 14.3  & 1.085E+04 & 0.652  & -4.905  & -1.803  & 0.402  & 1.252  & 3.7  \\
      & R101  & 4.137E+02 & 0.653  & -6.683  & -1.891  & 0.429  & 0.190  & 14.2  & 1.153E+04 & 0.613  & -4.889  & -2.000  & 0.392  & 1.204  & 3.8  \\
      \midrule
Deformable DETR & R50   & 4.226E+05 & 0.614  & -5.226  & -1.435  & 0.476  & 15.536  & 23.3  & 8.210E+04 & 0.660  & -4.327  & -2.108  & 0.425  & 1.307  & 2.8  \\
\midrule
Faster RCNN OI & R50   & -2.383E+00 & 0.614  & -6.401  & -2.036  & 0.434  & 3.037  & 20.2  & 1.880E+00 & 0.656  & -4.772  & -5.944  & 0.373  & 0.820  & 2.5  \\
RetinaNet OI & R50   & -8.290E-01 & 0.630  & -6.627  & -1.792  & 0.425  & 1.077  & 16.3  & -3.031E-01 & 0.655  & -4.784  & -2.507  & 0.375  & 1.105  & 3.1  \\
\midrule
SoCo  & R50   & -1.907E+01 & 0.612  & -6.568  & -1.656  & 0.427  & 3.065  & 20.6  & -2.707E+00 & 0.676  & -4.781  & -3.650  & 0.391  & 1.068  & 2.3  \\
InsLoc & R50   & -7.507E-02 & 0.540  & -6.625  & -0.925  & 0.417  & 0.690  & 18.8  & -1.149E-02 & 0.496  & -4.895  & 1.239  & 0.388  & 1.452  & 0.5  \\
\midrule
UP-DETR & R50   & -4.091E+02 & 0.428  & -6.851  & -1.341  & 0.345  & 0.062  & 13.5  & -5.645E+03 & 0.217  & -5.985  & -4.089  & 0.134  & 0.167  & 0.4  \\
DETReg & R50   & -1.594E+02 & 0.419  & -5.978  & -1.963  & 0.367  & 5.498  & 15.1  & -8.055E+03 & 0.411  & -5.119  & -4.957  & 0.272  & 0.562  & 2.0  \\
\midrule
\multicolumn{2}{l}{$\tau_w$} & 0.16  & 0.53  & 0.52  & 0.14  & \textbf{0.71 } & \textbf{0.71 } & N/A   & -0.14  & -0.30  & 0.13  & \textbf{0.61 } & -0.09  & 0.50  & N/A \\
\bottomrule
\end{tabular}%

    }
\end{table*}%

\section{More Experimental Results}
\paragraph{Ranking Performance.}
Except for Weighted Kendall’s tau ($\tau_w$) and Top-1 Relative Accuracy (Rel@1), we also evaluate the transferability metrics based on Weighted Pearson's coefficient ($\rho_w$) \cite{freedman2007statistics} and Recall@1 \cite{li2021ranking}, as shown in Table \ref{tab:appendix_pearson}.
Weighted Pearson's coefficient is used to measure the linear correlation between transferability scores and ground truth fine-tuning performance.
Recall@1 is used to measure the ratio of successfully selecting the model with best fine-tuning performance.
The evaluation is conducted on 1\% 33-choose-22 possible source model sets (over 1.9M).
Regarding $\rho_w$, we can draw the conclusion that Det-LogME outperforms all three SOTA methods consistently on $6$ downstream tasks by a large margin.
The IoU based metric IoU-LogME also performs well on $5$ datasets.
Regarding Recall@1, our proposed Det-LogME outperforms previous SOTA methods in average.

\paragraph{Detailed Ranking Results.}
\label{sec:appendix_moreresults}
We provide detailed raw ranking results of all $33$ pre-trained detectors on $6$ downstream tasks, including the transferability scores, ground truth performance (the average result of 3 runs with very light variance), and \emph{Weighted Kendall}'s tau $\tau_w$.
The results are provided in the following tables. Table \ref{tab:pascal_voc_cityscapes} shows results on Pascal VOC and CityScapes, Table \ref{tab:soda_crowdhuman} shows results on SODA and CrowdHuman, and Table \ref{tab:visdrone_deeplesion} contains results on VisDrone and DeepLesion.


{\small
\bibliographystyle{ieee_fullname}
\bibliography{wacv}
}

\end{document}